\documentclass{clv3}

\pdfoutput=1 

\usepackage{hyperref}

\usepackage{xcolor}
\definecolor{darkblue}{rgb}{0, 0, 0.5}
\hypersetup{colorlinks=true,citecolor=darkblue, linkcolor=darkblue, urlcolor=darkblue}

\usepackage{xurl} 

\issue{1}{1}{2021}

\historydates{Action Editor: Preslav Nakov. Submission received: 16 September 2021;
Accepted for publication: 6 December 2021.}

\dochead{}

\runningtitle{Ethics Sheet for Automatic Emotion Recognition}

\runningauthor{Saif M. Mohammad}

\begin{document}

\title{Ethics Sheet for Automatic Emotion Recognition and Sentiment Analysis}

\author{Saif M. Mohammad\thanks{National Research Council Canada. E-mail: saif.mohammad@nrc-cnrc.gc.ca.}}

\maketitle

\begin{abstract}
The importance and pervasiveness of emotions in our lives makes affective computing a tremendously important and vibrant line of work. Systems for automatic emotion recognition (AER) and sentiment analysis can be facilitators of enormous progress (e.g., in improving public health and commerce) but also enablers of great harm (e.g., for suppressing dissidents and manipulating voters). Thus, it is imperative that the affective computing community actively engage with the ethical ramifications of their creations. In this paper, I have synthesized and organized information from AI Ethics and Emotion Recognition literature to present fifty ethical considerations relevant to AER. Notably, the sheet fleshes out assumptions hidden in how AER is commonly framed, and in the choices often made regarding the data, method, and evaluation. Special attention is paid to the implications of AER  on privacy and social groups. Along the way, key recommendations are made for responsible AER. The objective of the sheet is to facilitate and encourage more thoughtfulness on why to automate, how to automate, and how to judge success \textbf{well before} the building of AER systems. Additionally, the sheet acts as a useful introductory document on emotion recognition  (complementing survey articles).
\end{abstract}


\section{Introduction}

Emotions play a central role in our lives. 
Thus affective computing, which deals with emotions and computation (often through AI systems) is a  tremendously important and vibrant line of work. It is a sweeping interdisciplinary area of study exploring both fundamental research questions (such as \textit{what are emotions?}) and commercial applications (such as \textit{can machines detect consumer sentiment?}).

In her seminal book, \textit{Affective Computing}, Dr.\@ Rosalind Picard described Automatic Emotion Recognition (AER) as: ``giving emotional abilities to computers” \cite{picard2000affective}.  
Such systems can be incredibly powerful:  facilitators of enormous progress, but also enablers of great harm. 
In fact, some of the recent commercial and governmental uses of emotion recognition have garnered considerable criticism, including:
infringing on one's privacy, exploiting vulnerable sub-populations, and 
also allegations of downright pseudo-science \cite{wakefield_2021,article19_2021,woensel_nevil_2019}.
Even putting aside high-profile controversies, emotion recognition impacts people and thus entails ethical considerations (big and small). 
Thus, it is imperative that the AER community actively engage with the ethical ramifications of their creations.

This article, which I refer to as an \textit{Ethics Sheet for AER}, is a critical reflection of this broad field of study with the aim of facilitating more responsible emotion research and appropriate use of the technology.
As described in \citet{mohammad2021ethics}, an Ethics Sheet for an AI Task is a semi-standardized document that
synthesizes and organizes information from AI Ethics and AI Task literature to
present a comprehensive array of ethical considerations for that task. 
Thus, in some ways, an ethics sheet is similar to survey articles, except here the focus is on ethical considerations.
It:\\[-20pt]
\begin{itemize}
    \item Fleshes out assumptions hidden in how the task is framed, and in the choices often made regarding the data, method, and evaluation.
    \vspace*{-1mm}
    \item Presents ethical considerations unique or especially relevant to the task.
    \vspace*{-1mm}
    \item Presents how common ethical considerations manifest in the task.
    \vspace*{-1mm}
    \item Presents relevant dimensions and choice points; along with tradeoffs. 
    \vspace*{-1mm}
    \item Lists common harm mitigation strategies.
    \vspace*{-1mm}
    \item Communicates societal implications of AI systems to researchers, developers, and the broader society.
\end{itemize}
\vspace*{-2mm}
\noindent The sheet should flesh out various ethical considerations that apply at the level of the task. It should also flesh out ethical consideration of common theories, methodologies, resources,  and practices used in building AI systems for the task. A good ethics sheet will question some of the assumptions that often go unsaid.\\[5pt]
\noindent \textbf{Primary motivation for creating an Ethics Sheet for AER:}
to provide a go-to point for a carefully compiled substantive engagement with the ethical issues relevant to emotion recognition; going beyond individual systems and datasets and drawing on knowledge from a large body of past work.
The document will be useful to anyone who wants to build or use emotion recognition systems/algorithms for research or commercial purposes. 
Specifically, the main benefits can be summarized by the list below:\\[-18pt]
\begin{enumerate}
     \item Encourages more thoughtfulness on why to automate, how to automate, and how to judge success 
      \textbf{well before} the building of AER systems.\\[-20pt]
    \item Helps us better navigate research and implementation choices.
    \vspace*{-1mm}
        \item Moves us towards consensus and  standards.
        \vspace*{-1mm}
    \item Helps in developing better post-production documents such as datasheets and model cards.
    \vspace*{-1mm}
    \item Has citations and pointers; acts as a jumping off point for further reading.
    \vspace*{-1mm}
    \item Helps engage the various stakeholders of an AI task with each other. Helps stakeholders challenge assumptions made by researchers and developers. Helps develop harm mitigation strategies.
    \vspace*{-1mm}
    \item Acts as a useful introductory document on emotion recognition (complements survey articles).
\end{enumerate}
\vspace*{-2mm}
\noindent Note that even though this sheet is focused on AER, many of the ethical considerations apply broadly to natural language tasks in general. Thus, it can serve as a useful template to build ethics sheets for other tasks.\\[5pt]
\noindent \textbf{Target audience:} The primary audience for this sheet are researchers, engineers, developers, and educators from various fields (especially NLP, ML, AI, data science, public health, psychology, and digital humanities) who build, make use of, or teach about AER technologies; however, much of the discussion should be accessible to various other stakeholders of AER as well, including 
policy/decision makers, and those who are impacted by AER.
I hope also 
that this sheet will act as a springboard for the
creation of a sheet where non-technical stakeholders are the primary audience.\\[5pt]
\noindent \textbf{Process:} My own research interests are at the intersection of emotions and language---to understand how we use language to express our feelings. I created this sheet to gather and organize my thoughts around responsible emotion recognition research, and hopefully it is of use to others as well. Discussions with various scholars from computer science, psychology, linguistics, neuroscience, and social sciences (and their comments on earlier drafts) have helped shape this sheet. An earlier draft of this material was also posted as a blog post with an explicit invitation for feedback. Valuable insights from the community were then incorporated into this document. 
That said, it should be noted that I do not speak for the AER community. There is no “objective” or ``correct” ethics sheet. This sheet should be taken as one perspective amongst many in the community. I welcome dissenting views and encourage further discussion. These can lead to periodically revised or new ethics sheets. As stated in \citet{mohammad2021ethics}:\\[-15pt]
\begin{quote}
\textit{Multiple ethics sheets can be created (by different teams and approaches) 
to reflect multiple perspectives, viewpoints, and what is important to different groups of people.
We should be wary of the world with single authoritative ethics sheets per task and no dissenting voices.}
\end{quote}

The rest of the paper is organized as follows: Section 2 is a preface to the ethics sheet, Section 3 presents the Ethics sheet for AER (50 considerations), and this is followed by summarizing thoughts in Section 4. 
The Appendix compiles a list of succinct recommendations for responsible AER (drawn from the discussions on ethical considerations in Section 3).




\section{Preface for the Ethics Sheet on AER}
Let us consider a few rapid-fire questions to set the context. A good ethics sheet makes us question our assumptions. So let us start at the top:\\[7pt]
\noindent \textbf{Q1. Should we be building AI systems for Automatic Emotion Recognition? Is it ethical to do so?}\\[2pt]
\noindent \textbf{A.} This is a good question. 
This sheet will not explicitly answer the question, but it will help in clarifying and thinking about it. This sheet will sometimes suggest that certain applications in certain contexts are good or bad ideas, but largely it will discuss what are the various considerations to be taken into account: whether to build or use a particular system, 
how to build or use a particular system,
what is more appropriate for a given context, how to assess success, etc.\\[2pt]
The above question is also somewhat under-specified. We first need to clarify...\\[7pt]
\noindent \textbf{Q2. What does automatic emotion recognition mean?}\\[3pt]
\noindent \textbf{A.} Emotion recognition can mean many things, and it has many forms. (This sheet will get into that.) Emotion recognition can be deployed in many contexts. For example, many will consider automated insurance premium decisions based on inferred emotions to be inappropriate. However, studying how people use language to express gratitude, sadness, etc.\@ is considered okay in many contexts. A human--computer interaction system benefits from being able to identify which utterances can convey anger, joy, sadness, hate, etc. (Not having such capabilities will lead to offensive, unempathetic, and inappropriate interactions.) Many other contexts are described in the sheet.\\[8pt]
\noindent \textbf{Q3. Can machines infer one’s true emotional state ever?}\\[2pt]
\noindent \textbf{A.} No. (This sheet will get into that.)\\[7pt]
\noindent \textbf{Q4. Can machines infer some small aspect of people’s emotions (or emotions that they are trying to convey) 
in some contexts, to the extent that it is *useful*?}\\[2pt]  
\noindent \textbf{A.} In my view, yes. In a limited way, this is analogous to machine translation or web search. The machine does not understand language, nor does it understand what the user really wants, nor the social, cultural, or embodied context, but it is able to produce a somewhat useful translation or search result with some likelihood; and it produces some amount of inappropriate and harmful results with some likelihood. However, unlike machine translation or search, emotions are much more personal, private, and complex.
People cannot fully determine each other's emotions. People cannot fully determine their own emotional state. But we make do with our limitations and infer emotions as best we can to function socially. 
We also have moral and ethical failures. We cause harm because of our limitations, and we harbor stereotypes and biases.

If machines are to be a part of this world and interact with people in any useful and respectful way, then they must have at least some limited emotion recognition capabilities; and thereby will also cause some amount of harm.
Thus, if we use them, it is important that we are aware of the limitations; design systems that protect and empower those without power; deploy them in the contexts they are designed for; use them to assist human decision making; and work to mitigate the harms they will perpetrate.
We need to hold AER systems to high standards, not just because it is a nice aspirational goal, but because machines impact people at scale (in ways that individuals rarely can) and emotions define who we are (in ways that other attributes rarely do).
I hope this sheet is useful in that regard.

\section{Main Sheet (version 1.0)}

This ethics sheet for Automatic Emotion Recognition has four sections: Modalities and Scope, Task, Applications, and Ethical Considerations. The first three are brief and set the context. The fourth presents various ethical considerations of AER as a numbered list, organized in thematic groups.

\subsection{Modalities and Scope}

\noindent \textbf{Modalities:} Work on AER has used a number of modalities (sources of input), including:\\[-20pt]
\begin{itemize}
    \item Facial expressions, gait, proprioceptive data (movement of body), gestures 
    \vspace*{-1mm}
    \item Skin and blood conductance, blood flow, respiration, infrared emanations
\vspace*{-1mm}
    \item Force of touch, haptic data (from sensors of force) 
    \vspace*{-1mm}
    \item Speech, language (esp. written text, emoticons, emojis) 
\end{itemize}
\vspace*{-3mm}
\noindent All of these modalities come with benefits, potential harms, and ethical considerations.\\[5pt]
\noindent \textbf{Scope:} This sheet will focus on AER from written text and AER in Natural Language Processing (NLP), but 
several of the listed considerations apply to AER in general (regardless of modality, and regardless of field such as NLP or Computer Vision).

\subsection{Task}
\label{sec:tasks}

Automatic Emotion Recognition (AER) from one’s utterances (written or spoken) is a broad umbrella term used to refer to a number of related tasks such as those listed below: (Note that each of these framings has ethical considerations and may be more or less appropriate for a given context.)\\[-20pt]
\begin{enumerate}
    \item Inferring emotions felt by the speaker (e.g., given Sara’s tweet, what is Sara feeling?); 
Inferring emotions of the speaker as perceived by the reader/listener (e.g., what does Li think Sara is feeling?); 
Inferring emotions that the speaker is attempting to convey (e.g., what emotion is Sara trying to convey?)
These may be correlated, but they can be different depending on the particular instance. The first framing “inferring emotions felt by the speaker” is fairly common in scientific literature, but also perhaps most often misused/misinterpreted. More on this in the ethical considerations section.
\vspace*{-1mm}
\item Inferring the intensity of the emotions discussed above.
\vspace*{-1mm}
\item Inferring patterns of speaker’s emotions over long periods of time, across many utterances; including the inference of moods, emotion dynamics, and emotional arcs (e.g., tracking character emotion arcs in novels and  tracking impact of health interventions on a patient’s well-being).
\vspace*{-1mm}
\item Inferring speaker’s emotions/attitudes/sentiment towards a target product, movie, person, idea, policy, entity, etc. (e.g., does Sara like the new phone?).
\vspace*{-1mm}
\item Inferring emotions evoked in the reader/listener (e.g., what feelings arise in Li on reading Sara’s tweet?).
This may be different among different readers because of their past experiences, personalities, and world-views: e.g., the same text may evoke different feelings among people with opposing views on an issue.
\vspace*{-1mm}
\item Inferring emotions of people mentioned in the text (e.g., given a tweet that mentions Moe, what emotional state of Moe is conveyed in the tweet?).
\vspace*{-1mm}
\item Inferring emotionality of language used in text (regardless of whose emotions) (e.g., is the tweet about happy things, angry feelings, etc.?).
\vspace*{-1mm}
\item Inferring how language is used to convey emotions such as joy, sadness, loneliness, hate, etc. 
\vspace*{-1mm}
\item Inferring the emotional impact of sarcasm, metaphor, idiomatic expression, dehumanizing utterance, hate speech, etc. 
\end{enumerate}
\vspace*{-2mm}
\noindent \textbf{Note 1:} The term Sentiment Analysis is commonly used to refer to the task described in bullet 4, especially in the context of product reviews (sentiment is commonly labeled as positive negative, or neutral) \cite{turney-2002-thumbs, pang-etal-2002-thumbs}. On the other hand, determining the predilection of a person towards a policy, party, issue, etc. is usually referred to as Stance Detection, and involves classes such as favour and against \cite{StanceSemEval2016,MohammadSK17}.\\[3pt]
\noindent \textbf{Note 2:} Many AER systems focus only on the emotionality of the language used (bullet 7), even though their stated goal might be one of the other bullets. This may be appropriate in restricted contexts such as customer reviews or personal diary blog posts, but not always. (More on this in \S\ref{sec:taskdesign} \textit{Ethical Considerations: Task Design}.)\\[3pt]
\noindent \textbf{Note 3:} There also exist tasks that focus not directly on emotions, but on associated phenomena, such as: whose emotions, who/what evoked the emotion, what types of human need was met or not met resulting in the emotion, etc.

See these surveys for more details: \citet{mohammad2020survey} examines emotions, sentiment, stance, etc.; \citet{zhang2018deep} focuses on sentiment analysis tasks; \citet{soleymani2017survey} surveys multi-modal techniques for sentiment analysis.

\subsection{Applications}
\label{sec:applications}

The potential benefits of AER are substantial.
Below is a sample of some existing applications: (Note that this is not an endorsement of these applications. All of the applications come with potential harms and ethical considerations. Use of AER by the military, for intelligence, and for education are especially controversial.)\\[-20pt]
\begin{itemize}
\item Public Health: Assist public health research projects, including those on loneliness \cite{guntuku2019studying,kiritchenko-etal-2020-solo}, depression \cite{de2013predicting,resnik-etal-2015-beyond}, suicidality prediction \cite{macavaney-etal-2021-community}, bipolar disorder \cite{karam2014ecologically}, stress \cite{eichstaedt2015psychological}, and well-being \cite{schwartz2013characterizing}. 
   \vspace*{-1mm} 
\item Commerce/Business: Track sentiment and emotions towards one’s products, track reviews, blog posts, YouTube videos and comments; develop virtual assistants, writing assistants; help advertise products that one is more likely to be interested in.
   \vspace*{-1mm} 
\item Government Policy and Public Health Policy: Tracking and documenting views of the broader public on a range of issues that impact policy (tracking amount of support and opposition, identifying underlying issues and pain points, etc.). Governments and health organizations around the world are also interested in tracking how effective their messaging has been in response to crises such as pandemics and climate change.
   \vspace*{-1mm} 
\item Art and Literature: Improve our understanding of what makes a compelling story, how do different types of characters interact, what are the emotional arcs of stories,
what is the emotional signature of different genres, 
what makes well-rounded characters, 
why does art evoke emotions, 
how do the lyrics and music impact us emotionally, etc. Can machines generate art  (generate paintings, stories, music, etc.)?
   \vspace*{-1mm} 
\item Social Sciences, Neuroscience, Psychology: Help answer questions about people. What makes people thrive? What makes us happy? What can our language tell us about our well-being? What can language tell us about how we construct emotions in our minds? How do we express emotions? How different are people in terms of what different emotion words mean to them and how they use emotional words? 
   \vspace*{-1mm} 
\item Military, Policing, and Intelligence: Tracking how sets of people or countries feel about a government or other entities (controversial); tracking misinformation on social media. 
\end{itemize}

\subsection{ETHICAL CONSIDERATIONS}
The usual approach to building an AER system is to \textbf{design the task} (identify the process to be automated, the emotions of interest, etc.), compile appropriate \textbf{data} (label some of the data for emotions—a process referred to as human annotation), train ML models that capture patterns of emotional expression from the data—\textbf{the method}, and \textbf{evaluate} the models by examining their predictions on a held-out test set.
There are ethical considerations associated with each step of this development process.
Considerations for privacy and social groups are especially pertinent for AER and  cut across task, design, data, and evaluation.

This section describes fifty considerations grouped under the themes: \textit{Task Design, Data, Method, Impact and Evaluation}, and \textit{Implications for Privacy and Social Groups}.  
First I present an outline of the considerations along with a summary for each grouping. This is followed by five sub-sections (\S\ref{sec:taskdesign} through \S\ref{sec:privacysg}) that present, in detail, the ethical considerations associated with the five groups.

\begin{quote}
    
\noindent \textbf{I. TASK DESIGN}\\[5pt]
\noindent Summary: This section discusses various ethical considerations associated with the choices involved in the framing of the emotion task and the implications of automating the chosen task. Some important considerations include: Whether it is even possible to determine one’s internal mental state? Whether it is ethical to determine such a private state? And, who is often left out in the design of existing AER systems? I discuss how it is important to consider which formulation of emotions is appropriate for a specific task/project; while avoiding careless endorsement of theories that suggest a mapping of external appearances to inner mental states.\\[5pt]
\noindent A. Theoretical Foundations\\[3pt]
\noindent \hspace*{3mm}   1. Emotion Task  and Framing\\
\hspace*{3mm}  2. Emotion Model and Choice of Emotions\\
\hspace*{3mm} 3. Meaning and Extra-Linguistic Information\\
\hspace*{3mm} 4. Wellness and Emotion\\
\hspace*{3mm} 5. Aggregate Level vs. Individual Level\\[3pt]
B. Implications of Automation\\[3pt]
\hspace*{3mm}  6. Why Automate (Who Benefits; Will this Shift Power)\\ 
\hspace*{3mm} 7. Embracing Neurodiversity\\
\hspace*{3mm} 8. Participatory/Emancipatory Design\\
\hspace*{3mm}  9. Applications, Dual use, Misuse\\
\hspace*{1.5mm}   10. Disclosure of Automation\\[3pt]
 
\noindent \textbf{II. DATA}\\[5pt]
\noindent Summary: This section has three 
themes: implications of using datasets of different kinds, the tension between human variability and machine normativeness, and the  considerations regarding the people who have produced the data. Notably, I discuss how on the one hand  is the tremendous variability in human mental representation and expression of emotions, and on the other hand, is the inherent bias of modern machine learning approaches to ignore variability. Thus, through their behaviour (e.g., by recognizing some forms of emotion expression and not  others), AI systems convey to the user what is ``normal''; implicitly invalidating other forms of emotion expression.\\[5pt]
C. Why This Data\\[3pt]
\hspace*{3mm}   11. Types of data\\
\hspace*{3mm}   12. Dimensions of data\\[3pt]
D. Human Variability vs.\@ Machine Normativeness\\[3pt]
\hspace*{3mm}  13. Variability of Expression and Mental Representation\\
\hspace*{3mm}  14. Norms of Emotions Expression\\
\hspace*{3mm}  15. Norms of Attitudes\\
 \hspace*{3mm} 16. One "Right" Label or Many Appropriate Labels\\
 \hspace*{3mm} 17. Label Aggregation\\
\hspace*{3mm}  18. Historical Data (Who is Missing and What are the Biases)\\
\hspace*{3mm}  19. Training-Deployment Differences\\[3pt]
E. The People Behind the Data\\[3pt]
\hspace*{3mm}  20. Platform Terms of Service\\
\hspace*{3mm}  21. Anonymization and Ability to Delete One's information\\
\hspace*{3mm}  22. Warnings and Recourse\\
\hspace*{3mm}  23. Crowdsourcing\\[3pt]
 
\noindent \textbf{III. METHOD}\\[5pt]
Summary: This section discusses the ethical implications of doing AER using a given method. It presents the types of methods and their tradeoffs, as well as, considerations of who is left out, spurious correlations, and the role of context. Special attention is paid to green AI and the fine line between emotion management and manipulation.\\[5pt]
F. Why This Method\\[3pt]
\hspace*{3mm}  24. Types of Methods and their Tradeoffs\\
\hspace*{3mm}  25. Who is Left Out by this Method\\
\hspace*{3mm}  26. Spurious Correlations\\
\hspace*{3mm}  27. Context is Everything\\
\hspace*{3mm}  28. Individual Emotion Dynamics\\
\hspace*{3mm}  29. Historical Behavior is not always indicative of Future Behavior\\
\hspace*{3mm}  30. Emotion Management, Manipulation\\
\hspace*{3mm}  31. Green AI\\[3pt]
 
\noindent \textbf{IV. IMPACT AND EVALUATION}\\[5pt]
Summary: This section discusses ethical considerations associated with the impact of AER systems using both traditional metrics as well as through a number of other criteria beyond metrics. Notably, this latter subsection discusses interpretability, visualizations, building safeguards, and contestability, because even when systems work as designed, there will be some negative consequences. Recognizing and planning for such outcomes is part of responsible development.\\[5pt]
G. Metrics\\[3pt]
\hspace*{3mm}  32. Reliability/Accuracy\\
\hspace*{3mm}  33. Demographic Biases\\
\hspace*{3mm}  34. Sensitive Applications\\
\hspace*{3mm}  35. Testing (on Diverse Datasets, on Diverse Metrics)\\[5pt]
H. Beyond Metrics\\[3pt]
\hspace*{3mm}  36. Interpretability, Explainability\\
\hspace*{3mm}  37. Visualization\\
\hspace*{3mm}  38. Safeguards and Guard Rails\\
\hspace*{3mm}  39. Harms even when the System Works as Designed\\
\hspace*{3mm}  40. Contestability and Recourse\\
\hspace*{3mm}  41. Be wary of Ethics Washing\\[3pt]

\noindent \textbf{V. IMPLICATIONS FOR PRIVACY, SOCIAL GROUPS}\\[5pt]
Summary: This section presents ethical implications of AER for privacy and for social groups. These issues cut across Task Design, Data, Method, and Impact. 
I discuss both individual and group privacy. 
The latter becomes especially important in the context of soft-biometrics determined through AER that are not intended to be able to identify individuals, but rather identify groups of people with similar characteristics. 
I discuss the need for work that does not treat people as a homogeneous group (ignoring sub-group differences) 
but rather explores disaggregation and  intersectionality, while minimizing reification and essentialization of social constructs.\\[5pt] 
I. Implications for Privacy\\[3pt]
\hspace*{3mm}  42. Privacy and Personal Control\\
\hspace*{3mm}  43. Group Privacy and Soft Biometrics\\
\hspace*{3mm}  44. Mass Surveillance vs. Right to Privacy, Expression, Protest\\
\hspace*{3mm}  45. Right Against Self-Incrimination\\
\hspace*{3mm}  46. Right to Non-Discrimination\\[5pt]
J. Implications for Social Groups\\[3pt]
\hspace*{3mm}  47. Disaggregation\\
\hspace*{3mm}  48. Intersectionality\\
\hspace*{3mm}  49. Reification and Essentialization\\
\hspace*{3mm}  50. Attributing People to Social Groups
 \end{quote}
\vspace*{-3mm}
\noindent One can read these various sections in one go, or simply use it as a reference when needed (jumping to sections of interest).

\subsubsection{TASK DESIGN} (Ten considerations.) \\[10pt]
\label{sec:taskdesign}
%
\noindent A. \textbf{Theoretical Foundations}\\[2pt]
\textit{Domain naivete is not a virtue.}\\[5pt]
Study the theoretical foundations for the task from relevant research fields such as psychology, linguistics, and sociology, to inform the  task formulation.\\[10pt]
\noindent \textbf{\#1. Emotion Task and Framing:} Carefully consider what emotion task should be the focus of the work (whether conducting human-annotation or building an automatic system). 
(See \S\ref{sec:tasks} for a sample of common emotion tasks.)
When building an AER system, a clear grasp of the task will help in making appropriate design choices. 
When choosing which AER system to use, a clear grasp of the emotion task most appropriate for the deployment context will help in choosing the right AER system.
It is not uncommon for users of AER to have a particular emotion task in mind and mistakenly assume that an off-the-shelf AER system is designed for that task.

Each of the emotion tasks has associated ethical considerations. For example,\\[-20pt]
\begin{quote}
\textit{Is the goal to infer one’s true emotions? Is it possible to comprehensively determine one’s internal mental state by any AI or human? (Hint: No.)
Is it ethical to determine such a private state?}
\end{quote}
\vspace*{-2mm}
\noindent Realize that it is impossible to capture the full emotional experience of a person (even if one had access to all the electrical signals in the brain).
A less ambitious goal is to infer some aspects of one’s emotional state.

Here, we see a distinct difference between AER that uses vision and AER that uses language. While there is little credible evidence of the connection between one’s facial expressions and one’s internal emotional state, there is a substantial amount of work on the idea that language is a window into one’s mind \cite{chomsky1975reflections,lakoff2008women,pinker2007stuff}---which of course also includes emotions \cite{bamberg1997language,wiebe2005annotating,tausczik2010psychological}.

That said, 
there is no evidence that
one can determine the full (or even substantial portions) of one's emotional state through their language. (See also considerations \#2 \textit{Emotion Model} and \#13 \textit{Variability of Expression} ahead on complexity of the emotional experience and variability of expression.)
Thus, often it is more appropriate to frame the AER task differently, for example, the objective could be:\\[-20pt]
\begin{itemize}
    \item to study how people express emotions: Work that uses speaker-annotated labeled data such as emotion-word hashtags in tweets usually captures how people convey emotions \cite{mohammad-2012-emotional,purver2012experimenting}. What people convey may not necessarily indicate what they feel.
   \vspace*{-1mm} 
\item to determine perceived emotion (how others may think one is feeling): 
Perceived emotions are not necessarily the emotions of the speaker.
Emotion annotations by people who have not written the source text usually reveal perceived emotions. (This is most common in NLP data-annotation projects.) Annotation aggregation strategies, such as majority voting usually only convey emotions perceived by a majority group. Are we missing out on the perceptions of some groups?  (More on majority voting in \#17 \textit{Label Aggregation}.)
   \vspace*{-1mm} 
\item to determine emotionality of language used in text (regardless of whose emotions, target/stimulus, etc.): This may be appropriate in some restricted-domain scenarios, for example, when one is looking at customer reviews. Here, the context is indicative that the emotionality in the language likely indicates attitude towards the product being reviewed. However, such systems have difficulty when dealing with movie and book reviews because then it has to distinguish between text expressing attitudes towards the book/movie from text describing what happened in the plot (which is likely emotional too).
\vspace*{-1mm}
\item to determine trends at aggregate level: Emotionality of language is also useful when tracking broad patterns at an aggregate level e.g., tracking trends of emotionality in tens of thousands of tweets or text in novels over time (e.g., \citet{paul2011you,mohammad-2011-upon,10.1145/2145204.2145347}). The idea is that aggregating information from a large number of instances leads to the determination of meaningful trends in emotionality. (See also discussion in \#5 \textit{Aggregate Level vs.\@ Individual Level}.)
\end{itemize}
\vspace*{-3mm}
\noindent In summary, it is important to identify what emotion task is the focus of one's work, use appropriate data, and communicate the nuance of what is being captured to the stakeholders. Not doing so will lead to the misuse and misinterpretation of one’s work.
Specifically, \textbf{AER systems should not claim to determine one's emotional state from their utterance, facial expression, gait, etc}. At best, AER systems capture what one is trying to convey or what is perceived by the listener/viewer, and even there, given the complexity of human expression, they are often inaccurate.
A separate question is whether AER systems can determine trends in the emotional state of a person (or a group) over time? Here, inferences are drawn at aggregate level from much larger amounts of data. Studies on public health, such as those listed in \S3.3 fall in this category. Here too, it is best to be cautious in making claims about mental state, and use AER as one source of evidence amongst many (and involve expertise from public health and psychology).\\

\noindent \textbf{\#2. Emotion Model and Choice of Emotions:} 
\noindent Work on AER needs to opertationalize the aspect of emotion it intends to capture, that is, decide on emotion-related categories or dimensions of interest, decide on how to represent them, etc.
Psychologists and neuro-scientists have identified several theories of emotion to inform these decisions:\\[-18pt]
\begin{itemize}
    \item The Basic Emotions Theory (BET): Work by Dr. Paul Ekman in 1960s galvanized the idea that some emotions (such as joy, sadness, fear, etc.) are universally expressed through similar facial expressions, and these emotions are more basic than others \cite{ekman1992there,ekman1994nature}. This was followed by other proposals of basic emotions by Robert Plutchik, Izard and others. However, many of the tenets of BET, such as the universality of some emotions and their fixed mapping to facial expressions, stand discredited or are in question \cite{barrett2017emotions,barrett2019emotional}.
\item The Dimensional Theory: Several influential studies have shown that the three most fundamental, largely independent, dimensions of affect and connotative meaning are valence (positiveness--negativeness / pleasure--displeasure), arousal (active--sluggish), and dominance (dominant-–submissive / in control--out of control) \cite{osgood1957measurement,russell1980circumplex,russell1977evidence,russell2003core}. Valence and arousal specifically are commonly studied in a number of psychological and neuro-cognitive explorations of emotion.
\item Cognitive Appraisal Theory: The core idea behind appraisal theory \cite{scherer1999appraisal,lazarus1991progress} is that emotions arise from a person’s evaluation of a situation or event. (Some varieties of the theory point to a parallel process of reacting to perceptual stimuli as well.) Thus it naturally accounts for variability in emotional reaction to the same event since different people may appraise the situation differently. Criticisms of appraisal theory centre around questions such as: whether emotions can arise without appraisal; whether emotions can arise without physiological arousal; and whether our emotions inform our evaluations.
\item The Theory of Constructed Emotions: Dr. Lisa Barrett proposed a new theory on how the human brain constructs emotions from our experiences of the world around us and the signals from our body \cite{barrett2017theory}.
\end{itemize}
\vspace*{-3mm}
\noindent Since ML approaches rely on human-annotated data (which can be hard to obtain in large quantities), AER research has often gravitated to the Basic Emotions Theory, as that work allows one to focus on a small number of emotions. This attraction has been even stronger in the vision AER research because of BET’s suggested mapping between facial expressions and emotions. However, as noted above, many of the tenets of BET stand debunked.

Consider which formulation of emotions is appropriate for your task/project. For example, one may choose to work with the dimensional model or the model of constructed emotions if the goal is to infer behavioural or health outcome predictions. 
Despite criticisms of BET, it makes sense for some NLP work to focus on \textit{categorical emotions} such as joy, sadness, guilt, pride, fear, etc. (including what some refer to as basic emotions) because people often talk about their emotions in terms of these concepts. Most human languages have words for these concepts (even if our individual mental representations for these concepts vary to some extent). However, note that work on categorical emotions by itself is not an endorsement of the BET. Do not refer to some emotions as basic emotions, unless you mean to convey your belief in the BET. Careless endorsement of theories can lead to the perpetuation of ideas that are actively harmful (such as suggesting we can determine internal state from outward appearance---physiognomy).\\


\noindent \textbf{\#3. Meaning and Extra-Linguistic Information:} The meaning of an utterance is not only a property of language, but it is grounded in human activity, social interactions, beliefs, culture, and other extra-linguistic events, perceptions, and knowledge 
\cite{harris1954distributional,chomsky2014aspects, ervin1973some,
bisk-etal-2020-experience, bender-koller-2020-climbing,hovy-yang-2021-importance}.
Thus one can express the same emotion in different ways in different contexts, different people express the same emotions in different ways, and the same utterances can evoke different emotions in different people. AER systems that do not take extra-linguistic information into consideration will always be limited in their capabilities, and risk being systematically biased, insensitive, and discriminatory. More on this in \#13 \textit{Variability of Expression} and \#14 \textit{Norms of Emotion Expression}.\\

\noindent \textbf{\#4. Wellness and Emotion:} The prominent role of one’s body in the theory of constructed emotion \cite{barrett2017emotions}, nicely accounts for the fact that various physical and mental illnesses (e.g., Parkinsons, Alzheimers, Cardiovascular Disease, Depression, Anxiety) impact our emotional lives. Existing AER systems are not capable of handling these inter-subject and within-subject variability and thus should not be deployed in scenarios where their decisions could negatively impact the lives of people; and where deployed, their limitations should be clearly communicated.

Emotion recognition is playing a greater role than ever before in understanding how our language reflects our wellness, 
understanding how certain physical and mental illnesses impact our emotional expression, and
understanding how emotional expression can help improve our well-being.
For some medical conditions, clinicians can benefit from a detailed history of one's emotional state. 
However, people are generally not very good at remembering how they had been feeling over the past week, month, etc.
Thus an area of interest is to use AER to help patients track their emotional state.
See applications of AER in Public Health in Section \ref{sec:applications}. See also CL Psych workshop proceedings. 
Note, however, that these are cases where the technology is working firmly in an assistive role to clinicians and psychologists---providing additional information in situations where human experts make decisions based on a number of other sources of information as well. See \citet{10.1145/3287560.3287587} for ethical considerations on inferring mental health states from one’s utterances.\\

\noindent \textbf{\#5. Aggregate Level vs. Individual Level:} Emotion detection can be be used to make inferences about individuals or groups of people; 
for example, to assist one in writing, to recommend products or services, etc.\@ or to determine broad trends in attitudes towards a product, issue, or some other entity. 
Statistical inferences tend to be more reliable when using large amounts of data and when using more relevant data. Systems that make predictions about individuals often have very little pertinent information about the individual and thus often fall back on data from groups of people. Thus, given the person-to-person variability and within-person variability discussed in the earlier bullets, systems are imbued with errors and biases.
Further, 
these errors are especially detrimental because of the direct and personal nature of such interactions. They may, for example, attribute majority group behavior/preferences to the individual, further marginalizing those that are not in the majority.\\[-20pt]
\begin{quote}
    \textit{Various ethical concerns, including privacy, manipulation, bias, and free speech, are further exacerbated when systems act on individuals.}
\end{quote}
\vspace*{-3mm}
\noindent Work on finding trends in large groups of people on the other hand benefits from having a large amount of relevant information to draw on. However, see \#43 \textit{Group Privacy} and \#47 to \#50 \textit{Implications for Social Groups}  for relevant concerns.\\[3pt]

\noindent \textbf{B. Implications of Automation}\\[2pt]
\textit{What are the ethical implications of automating the chosen task?}\\[5pt]

\noindent \textbf{\#6. Why Automate (Who Benefits and Will this Shift Power):} When we choose to work on a particular AER task, or any AI task for that matter, it is important to ask ourselves why? Often the first set of responses may be straightforward: e.g., to automate some process to make people’s lives easier, or to provide access to some information that is otherwise hard to obtain, or to answer research questions about how emotions work. However, lately there has been a call to go beyond this initial set of responses and ask more nuanced, difficult, and uncomfortable questions such as:\\[-20pt]
\begin{itemize}
    \item Who will benefit from this work and who will not \cite{trewin2019considerations}?
    \item Will this work shift power from those who already have a lot of power to those that have less power \cite{kalluri2020don}?
   \item How can we reframe or redesign the task so that it helps those that are most in need \cite{monteiro2019ruined}?
\end{itemize}
\vspace*{-3mm}


\noindent Specifically for AER, this will involve considerations such as:\\[-20pt]
\begin{itemize}
    \item Are there particular groups of people who will not benefit from this task: e.g., people who convey and detect emotions differently than what is common (e.g., people on the autism spectrum), people who use language differently than the people whose data is being used to build the system (e.g., older people or people from a different region)?
    \item If AER is used in some application, say to determine insurance premiums, then is this further marginalizing those that are already marginalized?
    \item How can we prevent the use of emotion and stance detection systems for detecting and suppressing dissidents?
    \item How can AER help those that need the most help?
    \vspace*{-3mm}
\end{itemize}
\noindent Various other considerations such as those listed in this sheet can be used to further evaluate the wisdom in investing our labor in a particular task.\\

\noindent \textbf{\#7. Embracing Neurodiversity:} Much of the ML/NLP emotion work has assumed homogeneity of users and ignored neurodiversity, alexithymia, and autism spectrum. 
These groups have significant overlap, but are not identical. They are also often characterized as having difficulty in sensing and expressing emotions. Therefore these groups hold particular significance in the development of an inclusive AER system. Existing AER systems implicitly cater to the more populous neurotypical group. At minimum, such AER systems should explicitly acknowledge this limitation. Report disaggregated performance metrics for relevant groups. (See also \#47 \textit{Disaggregation}.)

Greater research attention needs to be paid to the neurodiverse group. When doing data annotations, we should try to obtain information on whether participants are neurodiverse or neurotypical (when participants are comfortable sharing that information), and include that information at an aggregate level when we report participant demographics. Work in Psychology has used scales such as the Toronto Alexithymia Scale (TAS-20) to determine the difficulty that people might have in identifying and describing emotions \cite{bagby1994twenty}.\\ 

\noindent \textbf{\#8. Participatory/Emancipatory Design:} Participatory design in research and systems development centers the people, especially marginalized and disadvantaged communities, such that they are not mere passive subjects but rather have the agency to shape the design process \cite{spinuzzi2005methodology}. This has also been referred to as emancipatory research \cite{humphries2020arguments, noel2016promoting, oliver1997emancipatory} 
and is pithily captured by the rallying cry “nothing about us without us”. These calls have developed across many different domains, including research pertaining to disability \cite{stone1996parasites,seale2015negotiating}, 
indigenous communities \cite{hall2014not}, autism spectrum 
\cite{fletcher2019making,bertilsdotter2019doing}, and neurodiversity 
\cite{brosnan2017beyond,10.1007/978-3-030-25629-6_42}.
See \citet{10.1007/978-3-030-25629-6_42} for specific recommendations for conducting studies with neuro-diverse participants.\\

\noindent \textbf{\#9. Applications, Dual Use, Misuse:} AER is a powerful enabling technology that has a number of applications. Thus, like all enabling technologies it can be misused and abused.
Examples of inappropriate commercial AER application include:\\[-20pt]
\begin{itemize}
    \item Using AER at airports to determine whether an individual is dangerous simply from their facial expressions.
    \vspace*{-1mm}
    \item Detecting stance towards governing authorities to persecute dissidents.
    \vspace*{-1mm}
    \item Using deception detection or lie detection en masse without proper warrants or judicial approval. (Using such technologies even in carefully restricted individual cases is controversial.)
    \vspace*{-1mm}
    \item Increasing someone’s insurance premium because the system has analyzed one’s social media posts to determine (accurately or inaccurately) that they are likely to have a certain mental health condition.
    \vspace*{-1mm}
    \item Advertisement that prey on the emotional state of people, e.g., user-specific advertising to people when they are emotionally vulnerable.
\end{itemize}
\vspace*{-1mm}
\noindent \textit{Socio-Psychological Applications:} Applications such as inferring patterns in emotions of a speaker to in turn infer other characteristics such as suitability for a job, personality traits, or health conditions are especially fraught with ethical concerns. For example, consider the use of the Myers–Briggs Type Indicator (MBTI) for hiring decisions or 
research on detecting personality traits automatically. Notable ethical concerns, include:\\[-20pt]
\begin{itemize}
    \item MBTI is criticized by psychologists, especially for its lack of test-retest reliability \cite{Boyle95,gerras2016moving,grant_2013}. The Big 5 personality traits formalism \cite{COBBCLARK201211} has greater validity, but even when using Big 5, it is easy to overstate the conclusions.
    \vspace*{-1mm}
    \item Even with accurate personality trait identification, there is little to no evidence that using personality traits for hiring and team-composition decisions is beneficial. The use of such tests have also been criticized on the grounds of discrimination \cite{snow_2020}.
\end{itemize}
\vspace*{-1mm}

\noindent \textit{Health and Well-Being Applications:} AER has considerable potential for improving our health and well-being outcomes. However, the sensitive nature of such applications require substantial efforts to adhere to the best ethical principles. For example, how can harm be mitigated when systems make errors? Should automatic systems be used at all given that sometimes we cannot put a value to the cost of errors? What should be done when the system detects that one is at a high risk of suicide, depression, or some other severe mental health condition? How to safeguard patient privacy? See the shared task at the 2021 CL Psych workshop where a secure enclave was used to store the training and test data. 
See these papers for ethical considerations of AI systems in health care 
\cite{yu2018artificial,lysaght2019ai,panesar2019machine}.\\

\noindent \textit{Applications in Art and Culture:} Lately there has been increasing use of AI in art and culture, especially through curation and recommendation systems. See \citet{Born21} for a discussion of ethical implications, including: are we really able to determine what art one would like, long-term impacts of automated curation (on users and artists), and diversity of sources and content.

AI is also used in the analysis and generation of art: e.g, for literary analysis and generating poems, paintings, songs, etc. Since emotions are a central component of art, much of this work also includes automatic emotion recognition: e.g. tracking the emotions of characters in novels, recommending songs for people based on their mood, and generating emotional music. This raises several questions including:\\[-20pt]
\begin{itemize}
    \item Is it art if the creation did not involve human input?\footnote{https://www.artbasel.com/news/artificial-intelligence-art-artist-boundary}
    \vspace*{-1mm}
    \item Should AI play a collaborative role with other artists (enhancing their creativity) as opposed to generate pieces on its own?
    \vspace*{-1mm}
    \item How will artists be impacted by AI’s role in art?
    \vspace*{-1mm}
    \item Who should get credit for AI art?\footnote{https://www.cnn.com/style/article/ai-art-who-should-get-credit-conversation/index.html}
    \vspace*{-1mm}
    \item     How should we critique AI art?\footnote{https://www.artnews.com/art-in-america/features/creative-ai-art-criticism-1202686003/}
\end{itemize}
\vspace*{-3mm}
\noindent See further discussion by \citet{hertzmann2020computers}.\\

\noindent \textbf{\#10. Disclosure of Automation:} Disclose to all stakeholders the decisions that are being made (in part or wholly) by automation. Provide mechanisms for the user to understand why relevant predictions were made, and also to contest the decisions. (See also \#36 \textit{Interpretability} and \#40 \textit{Contestability}.)

Artificial agents that perceive and convey emotions in a human-like manner can give one the impression that they are interacting with a human. Artificial agents should begin their interactions with humans by first disclosing that they are artificial agents \cite{dickson_2018}, even though some studies show certain negative outcomes of such a disclosure 
\cite{de2020should,mozafari2020chatbot}. \\


\subsubsection{DATA} (Thirteen considerations.)\\ [-2pt]
\label{sec:data}


\noindent \textbf{C. Why This Data}\\
\textit{What are the ethical implications of using the chosen data}?\\[5pt]

\noindent \textbf{\#11. Types of Data:} Emotion and sentiment researchers have used text data, speech data, data from mobile devices, data from social media, product reviews, suicide notes, essays, novels, movie screenplays, financial documents, etc. All of these entail their own ethical considerations in terms of the various points discussed in this article. AER systems use data in various forms, including:\\[-20pt]
\begin{itemize} 
    \item \textit{Large Language Models:} Language models such as BERT (that capture common patterns in language use) are obtained by training ML models on massive amounts of text found on the internet. See \citet{bender2021dangers} for ethical considerations in the use of large language models, including: documentation debt, difficult to curate, incorporation of inappropriate biases, and perpetuation of stereotypes. Note also that using smaller amounts of data raise concerns as well: they may not have enough generalizable information; they may be easier to overfit on; and they may not include diverse perspectives. An important aspect of preparing data (big or small) is deciding how to curate it 
    (e.g., what to discard).
    \item \textit{Emotion Lexicons:} Emotion Lexicons are lists of words and their associated emotions (determined manually by annotation or automatically from large corpora). Word--emotion association lexicons (such as AFINN \cite{nielsen2011new}, NRC Emotion Lexicon \cite{Mohammad13}, and the Valence, Arousal, Dominance Lexicon \cite{vad-acl2018}) are a popular type of resource used in emotion research, emotion-related data science, and machine learning models for AER. See \citet{mohammad2020practical} for biases and ethical considerations in the use of such emotion lexicons. Notable among these considerations is how words in different domains often convey different senses and thus have different emotion associations. Also, word associations capture historic perceptions that change with time and may differ across  different groups of people. They are not indicative of inherent immutable emotion labels.
    \item \textit{Labeled Training and Testing Data:} AER systems often make use of a relatively small number of example instances that are manually labeled (annotated) for emotions. A portion of these is used to train/fine-tune the large language model (training set). The rest is further split for development and testing. I discuss various ethical considerations associated with using emotion-labeled instances below.
\end{itemize}
\noindent \textbf{\#12. Dimensions of Data:} The data used by AER systems can be examined across various dimension: size of data; whether it is custom data (carefully produced for the research) or data obtained from an online platform (naturally occurring data); less private/sensitive data or more private/sensitive data; what languages are represented in the data; degree of documentation provided with the data; and so on. All of these have societal implications and the choice of datasets should be appropriate for the context of deployment.\\[3pt]

\noindent \textbf{D. Human Variability vs.\@ Machine Normativeness}\\[5pt]
\noindent \textit{What should we know about emotion data so that we use it appropriately?}\\

\noindent \textbf{\#13. Variability of Expression and Mental Representation:} Language is highly variable---we can express roughly the same meaning in many different ways.\\[-18pt]
\begin{quote}
\textit{Expressions of emotions through language are highly variable: Different people express the same emotion differently; the same text may convey different emotions to different people.
}
\end{quote}
\vspace*{-2mm}
\noindent This is true even for people living in the same area and especially true for people living in different regions, and people with different lived experiences.
Some cues of emotion are somewhat more common and somewhat more reliable than others. This is usually the signal that automatic systems attempt to capture.
We construct emotions in our brains from the signals we get from the world and the signal we get from our bodies. This mapping of signals to emotions is highly variable, and different people can have different signals associated with different emotions \cite{barrett2017theory}; therefore, different people have different concept--emotion associations. For example, high school, public speaking, and selfies may evoke different emotions in different people.
This variability is not to say that there are no commonalities. In fact, speakers of a language share substantial commonalities in their mental representation of concepts (including emotions), which enables them to communicate with each other. However,
the variability should also be taken into consideration when building datasets, systems, and choosing where to deploy the systems.\\


\noindent \textbf{\#14. Norms of Emotion Expression:}
As John M. Culkin once said, ``\textit{We shape our tools and thereafter they shape us.}"
Whether text, speech, vision, or any other modality, AI systems are often trained on a limited set of emotion expressions and their emotion annotations (emotion labels for the expressions).\\[-20pt]
\begin{quote}
\textit{Thus, through their behaviour (e.g., by recognizing some forms of emotion expression and not recognizing others), AI systems convey to the user that it is “normal” or appropriate to convey emotions in certain ways; implicitly invalidating other forms of emotion expression.}\\[-15pt]
\end{quote}
Therefore it is important for emotion recognition systems to accurately map a diverse set of emotion instantiations to emotion categories/dimensions. That said, it is also worth noting that the variations in emotion and language expression are so large that systems can likely never attain perfection. 
The goal is to obtain useful levels of emotion recognition capabilities without having systematic gaps that convey a strong sense of emotion-expression normativeness.

Normative implications of AER are analogous to normative implications of movies (especially animated ones):\\[-20pt]
\begin{itemize}
    \item Badly executed characters express emotions in fixed stereo-typical ways.
    \item Good movies explore the diversity, nuance, and subtlety of human emotion expression.
    \item Influential movies (bad and good) convey to a wide audience around the world how emotions are expressed or what is “normal” in terms of emotion expression. Thus they can either colonize other groups, reducing emotion expression diversity, or they can validate one's individualism and independence of self-expression.
\end{itemize}
\vspace*{-3mm}
Since AI systems are  influenced by the data they train on, dataset development should:\\[-20pt]
\begin{itemize}
    \item Obtain data from a diverse set of sources. Report details of the sources.
    \vspace*{-1mm}
    \item Studies have shown that a small percentage of speakers often produce a large percentage of utterances (see study by \cite{auxier2021social} on tweets). Thus, when creating emotion datasets, limit the number of instances included per person. 
    \citet{mohammad-kiritchenko-2018-understanding} kept one tweet for every query term and tweeter combination when studying relationships between affect categories (data also used in a SemEval-2018 Task 1 on emotions). 
    \citet{kiritchenko-etal-2020-solo} kept at most three tweets per tweeter when studying expressions of loneliness.
    \vspace*{-1mm}
    \item Obtain annotations from a diverse set of people. Report aggregate-level demographic information of the annotators.
\end{itemize}
\vspace*{-3mm}
\noindent Variability is common not just for emotions but also for  language. People convey meaning in many different ways. 
Thus, these considerations apply to NLP in general.\\

\noindent \textbf{\#15. Norms of Attitudes:} Different people and different groups of people might have different attitudes, perceptions, and associations with the same product, issue, person, social groups, etc. Annotation aggregation, by say majority vote, may convey a more homogeneous picture to the ML system. Annotation aggregation may also capture stereotypes and inappropriate associations for already marginalized groups. (For example, majority group A may perceive a minority group B as less competent, or less generous.) Such inappropriate biases are also encoded in large language models. When using language models or emotion datasets, assess the risk of such biases for the particular context and take correcting action as appropriate.\\

\noindent \textbf{\#16. One “Right” Label or Many Appropriate Labels:} When designing data annotation efforts, consider whether there is a “right” answer and a “wrong”? Who decides what is correct/appropriate? Are we including the voices of those that are marginalized and already under-represented in the data? 
When working with emotion and language data, there are usually no ``correct'' answers, but rather, some answers are more appropriate than others. And there can be multiple appropriate answers.\\[-20pt]
\begin{itemize}
\item If a task has clear correct and wrong answers and knowing the answers requires some training/qualifications, then one can employ domain experts to annotate the data. However, as mentioned, emotion annotations largely do not fall in this category.
\item If the goal is to determine how people use language, and there can be many appropriate answers, or we want to know how people perceive words, phrases, and sentences then we might want to employ a large number of annotators. This is much more in line with what is appropriate for emotion annotations — people are the best judges of their emotions and of the emotions they perceive from utterances.\\[-20pt]
\end{itemize}
\vspace*{-3mm}
\noindent Seek appropriate demographic information (respectfully and ethically).  Document annotator demographics, annotation instructions, and other relevant details. These are useful
in conveying to the reader that there is no one ``correct'' answer and that the dataset is situated in who annotated the data, the precise annotation instructions, when the data was annotated, etc.\\

\noindent \textbf{\#17. Label Aggregation:} Multiple annotations (by different people) for the same instance  are usually aggregated 
by choosing the majorty label. However, majority voting tends to capture majority group attitudes (at the expense of other groups). (See also \citet{aroyo2015truth}, \citet{checco2017let}, and \citet{klenner2020harmonization}.) As a result, sometimes researchers have released not just the aggregated results but also the raw (pre-aggregated data), as well as various versions of aggregated results. Others have argued in favor of not doing majority voting at all and including all annotations as input to ML systems \cite{basile2020s}. However, saying all voices should be included has its own problems: e.g., how to address and manage inappropriate/racist/sexist opinions; how to disentangle low-frequency valid opinions from genuine annotation errors and malicious annotations? (See also \#15 \textit{Norms of Attitudes} and \#47 \textit{Disaggregation}.)

If using majority voting, acknowledge its limitations. Acknowledge that it may be missing some/many voices.
Explore statistical approaches to finding multiple appropriate labels, while still discarding noise.
Employ separate manual checks to determine whether the human annotations also capture inappropriate human biases. Such biases may be useful for some projects (e.g., work studying such biases), but not for others. Warn users of  inappropriate biases that may exist in the data; and suggest strategies to deal with them when using the dataset.\\

\noindent \textbf{\#18. Historical Data (Who is Missing and What are the Biases):} Machine learning methods feed voraciously on data (often historical data). Natural language processing systems often feed on huge amounts of data collected from the internet. However, the data is not representative of everyone and seeped into this data are our biases.
Historical data over-represents people who have had power, who are more well to do, mostly from the west, mostly English-speaking, mostly white, mostly able-bodied, and so on and so forth. So the machines that feed on such data often learn their perspectives at the expense of the views of those already marginalized.

When using any dataset, devote resources to study who is included in the dataset and whose voices are missing. Take corrective action as appropriate.
Keep a portion of your funding for work with marginalized communities. Keep a portion of your funding for work on less-researched languages \cite{ruder_2020}.\\


\noindent \textbf{\#19. Training–Deployment Data Differences:} The accuracy of supervised systems is contingent on the assumption that the data the system is applied to is similar to the data the system was trained on. Deploying an off-the-shelf sentiment analysis system on data in a different domain, from a different time, or a different class distribution than the training data will likely result in poor predictions. Systems that are to be deployed to handle open-domain data should be trained on many diverse datasets and tested on many datasets that are quite different from the training datasets.\\[5pt]

\noindent \textbf{E. The People Behind the Data}\\[2pt]
\textit{What are the ethical implications on the people who have produced the data?}\\[4pt]
\noindent When building systems, we make extensive use of (raw and emotion-labeled) data. It can sometimes be easy to forget that behind the data are the people that produced it, and imprinted in it are a plethora of personal information.\\

\noindent \textbf{\#20. Platform Terms of Service:} Data for ML systems is often scraped from websites or extracted from large online platforms (e.g., Twitter, Reddit) using APIs. The terms of service for these platforms often include protections for the users and their data. Ensure that the terms of service of the source platforms are not violated: e.g., data scraping is allowed and data redistribution is allowed (in raw form or through ids). 
Ensure compliance with the robot exclusion protocol.\\ 

\noindent \textbf{\#21. Anonymization and Ability to Delete One’s information:} Take actions to anonymize data when dealing with private data; e.g., scrub identifying information. Some techniques are better at anonymization than others. (See for example, privacy-preserving work on word embeddings and sentiment data by \citet{thaine-penn-2021-chinese}.) Provide mechanisms for people to remove their data from the dataset if they choose to.\\[-20pt]
\begin{quote}
\textit{Choose to not work with a dataset if adequate safeguards cannot be placed.}
\end{quote}

\noindent \textbf{\#22. Warnings and Recourse:} Annotating highly emotional, offensive, or suicidal utterances can adversely impact the well-being of the annotators. Provide appropriate warnings. Minimize amount of data exposure per annotator. Provide options for psychological help as needed.\\

\noindent \textbf{\#23. Crowdsourcing:} Crowdsourcing (splitting a task into multiple independent units and uploading them on the internet so that people can solve them online) has grown to be a major source of labeled data in NLP, Computer Vision, and a number of other academic disciplines. Compensation often gets most of the attention when talking about crowdsourcing ethics, but there are several ethical considerations involved with such work such as: worker invisibility, lack of learning trajectory, humans-as-a-service paradigm, worker well-being, and worker rights. See
\citet{dolmaya2011ethics,fort-etal-2011-last,standing2018ethical,irani2013turkopticon,shmueli2021beyond}.
See (public) guidelines by AI2 for its researchers \cite{ai2_2019}.\\[-2pt]

\subsubsection{METHOD} (Eight considerations.)\\
\label{sec:method}


\noindent \textbf{F. Why This Method}\\[2pt]
\textit{What are the ethical implications of using a given method?}\\[-2pt]

\noindent \textbf{\#24. Methods and their Tradeoffs:} Different methods entail different trade-offs:\\[-20pt]
\begin{itemize}
    \item Less Accurate vs.\@ More Accurate: This usually gets all the attention; value other dimensions listed below as well. (See also \S \ref{sec:impact} IMPACT.)
    \vspace*{-1mm}
    \item White Box (can understand why system makes a given prediction) vs.\@ Black Box (do not know why it makes a given prediction): understanding the reasons behind a prediction help identify bugs and biases; helps contestability; arguably, better suited for answering research questions about language use and emotions.
    \vspace*{-1mm}
    \item Less Energy Efficient vs.\@ More Energy Efficient: See discussion further below on Green AI.
    \vspace*{-1mm}
    \item Less Data Hungry vs.\@ More Data Hungry: data may not always be abundant; needing too much data of a person leads to privacy concerns.
    \vspace*{-1mm}
    \item Less Privacy Preserving vs.\@ More Privacy Preserving: There is greater appreciation lately for the need for privacy-preserving NLP. 
    \vspace*{-1mm}
    \item Fewer Inappropriate Biases vs.\@ More Inappropriate Biases: We want our algorithms to not perpetuate/amplify inappropriate human biases.
\end{itemize}
\vspace*{-2mm}
\noindent Consider various dimensions of a method and their importance for the particular system deployment context before deciding on the method. Focusing on fewer dimensions may be okay in a research system, but widely deployed systems often require a good balance across the many dimensions.\\

\noindent \textbf{\#25. Who is Left Out:} The dominant paradigm in Machine Learning and NLP is to use use large pre-trained models pre-trained on massive amounts of raw data (unannotated text, pictures, videos, etc.) and then fine-tuned on small amounts of labeled data (e.g., sentences labeled with emotions) to learn how to perform a particular task. As such, these methods tend to work well for people that are well-represented in the data (raw and annotated), but not so well for others. (See also \#18 \textit{Historical Data}.)\\[-20pt]
\begin{quote}
\textit{Even just documenting who is left out is a valuable contribution.} 
\end{quote}
\vspace*{-3mm}
Explore alternative methods that are more inclusive, especially for those not usually included by other systems.\\

\noindent \textbf{\#26. Spurious Correlations:} Machine learning methods have been shown to be susceptible to spurious correlations. For example, 
\citet{agrawal2016analyzing} show that  when asked what is the ground covered with, visual QA systems tend to always say \textit{snow}, because in the training set, this question was only asked for when the ground was covered with snow. \citet{winkler2019association} and \citet{bissoto2020debiasing} show spurious correlations in melanoma and skin lesion detection systems. \citet{poliak-etal-2018-hypothesis} and \citet{gururangan2018annotation} show that natural language inference systems can sometimes decide on the prediction just from information in the premise, without regard for the hypothesis (for example, because a premise with negation is often a contradiction in the training set).

Similarly, machine learning systems capture spurious correlations when doing AER. For example, marking some countries and people of some demographics with less charitable and stereotypical sentiments and emotions. This phenomenon is especially marked in abusive language detection work where it was shown that data collection methods in combination with the ML algorithm result in the system marking any comment with identity terms such as gay, muslim, and jew as offensive.

Consider how the data collection and machine learning set ups can be addressed to avoid such spurious correlations, especially correlations that perpetuate racism, sexism, and stereotypes.
In extreme cases, spurious correlations lead to pseudoscience and physiognomy. For example, there have been a spate of papers attempting to determine criminality, personality, trustworthiness, and emotions just from one’s face or outer appearance. Note that sometimes, systematic idiosyncrasies of the data can lead to apparent good results on a held out test set even on such tasks. Thus it is important to consider whether the method and sources of information used are expected to capture the phenomenon of interest? Is there a risk that the use of this method may perpetuate false beliefs and stereotypes? If yes, take appropriate corrective action.\\

\noindent \textbf{\#27. Context is Everything:} Considering a greater amount of context is often crucial in correctly determining emotions/sentiment. What was said/written before and after the target utterance? Where was this said? What was the intonation and what was emphasized? Who said this? And so on. More context can be a double-edged sword though. The more the system wants to know about a person to make better predictions, the more we worry about privacy.
Work on determining the right balance between collecting more user information and privacy considerations, as appropriate for the context in which the system is deployed.\\

\noindent \textbf{\#28. Individual Emotion Dynamics:} A form of contextual information is one’s utterance emotion dynamics \cite{hollenstein2015time}. The idea is that different people might have different steady states in terms of where they tend to most commonly be (considering any affect dimension of choice). Some may move out of this steady state often, but some may venture out less often. 
Some recover quickly from the deviations, and for some it may take a lot of time.
Similar emotion dynamics occur in the text that people write or the words they utter—Utterance Emotion Dynamics \cite{hipson2021emotion}. The degree of correlation between the utterance emotion dynamics and the true emotion dynamics may be correlated, but one can argue that examining utterance emotion dynamics is valuable on its own.

Access to utterance emotion dynamics provides greater context and helps judge the degree of emotionality of new utterances by the person. Systems that make use of such detailed contextual information are more likely to make appropriate predictions for diverse groups of people. However, the degree of personal information they require warrants care, concern, and meaningful consent from the users.\\


\noindent \textbf{\#29. Historical behavior is not always indicative of future behavior (for groups and individuals):} Systems are often trained on static data from the past. However, perceptions, emotions, and behavior change with time. Thus automatic systems may make inappropriate predictions on current data. (See also \#18 \textit{Historical Data}.)\\[-1pt]


\noindent \textbf{\#30. Emotion Management, Manipulation:} Managing emotions is a central part of any human--computer interaction system (even if this is often not an explicitly stated goal). Just as in human--human interactions, we do not want the systems we build to cause undue stress, pain, or unpleasantness. For example, a chatbot has to be careful to not offend or hurt the feelings of the user with which it is interacting. For this, it needs to assess the emotions conveyed by the user, in order to then be able to articulate the appropriate information with appropriate affect.

However, this same technology can enable companies and governments to detect one’s emotions to manipulate their behavior. For example, it is known that we purchase more products when we are sad. So sensing when you are most susceptible to suggestion to plant ideas of what to buy, who to vote for, or who to dislike, can have dangerous implications. On the other hand, identifying how to cater to individual needs to improve their compliance with public health measures in a world-wide pandemic, or to help people give up on smoking, may be seen in more positive light. As with many things discussed in this article, consider the context to determine what levels of emotional management and meaningful consent are appropriate.\\[-1pt]

\noindent \textbf{\#31. Green AI:} A direct consequence of using ever-increasing pre-trained models (large number of training examples and hyperparameters) for AI tasks is that these systems are now drivers of substantial energy consumption. 
Recent papers showing the increasing carbon footprint of AI systems and approaches to address them \cite{Strubell_Ganesh_McCallum_2020,schwartz2020green}.
Thus, there is a growing push to develop AI methods that are not singularly focused on accuracy numbers on test sets, but are also mindful of efficiency and energy consumption \cite{schwartz2020green}. The authors encourage reporting of cost per example, size of training set, number of hyperparameters, and budget-accuracy curves. They also argue for regarding efficiency as a valued scientific contribution.

\subsubsection{IMPACT AND EVALUATION} (Ten considerations.)\\
\label{sec:impact}


\noindent \textbf{G. Metrics}\\ [2pt]
\noindent \textit{All evaluation metrics are misleading. Some metrics are more useful than others.}\\[-2pt]

\noindent \textbf{\#32. Reliability/Accuracy:} No emotion recognition method is perfect. However, some approaches are much less accurate than others. 
Some techniques are so unreliable that they are essentially pseudoscience. For example, trying to predict personality, mood, or emotions through physical appearances has long been criticized \cite{physiognomy_2017}. 
The ethics of a number of existing commercial systems that purportedly detect emotions from facial expressions is called into question
 by \citet{barrett2019emotional}, which shows the low reliability of recognizing emotions from facial expressions.\\


\noindent \textbf{\#33. Demographic Biases:} Some systems can be unreliable or systematically inaccurate for certain groups of people, races, genders, people with health conditions, people that are on the autism spectrum, people from different countries, etc. Such systematic errors can occur when working on:\\[-20pt]
\begin{itemize}
    \item Utterances of a group or faces of a group: For example, low accuracy in recognizing emotions in text produced by African Americans or in recognizing faces of African Americans \cite{buolamwini2018gender}.
    \item Utterances mentioning a group: For example, systematically marking texts mentioning African Americans as more angry, or texts mentioning women as more emotional \cite{kiritchenko-mohammad-2018-examining}.
\end{itemize}
\vspace*{-3mm}
\noindent Determine and present disaggregated accuracies. Take steps to address disparities in performance across groups. (See also \#47 \textit{Disaggregation}.)\\

\noindent \textbf{\#34. Sensitive Applications:} Some applications are considerably more sensitive than others and thus necessitate the use of a much higher quality of emotion recognition systems (if used at all). 
Automatic systems may sometimes be used in high-stakes applications if their role is to assist human experts. For example, assisting patients and health experts in tracking the patient’s emotional state.\\

\noindent \textbf{\#35. Testing (on Diverse Datasets, on Diverse Metrics):} Results on any test set are contingents on the attributes of that test set and may not be indicative of real-world performance, or implicit biases, or systematic errors of many kinds. Good practice is to test the system on many different datasets that explore various input characteristics. For example, see these evaluations that cater to a diverse set of emotion-related tasks, datasets, linguistic phenomena, and languages: SemEval 2014 Task 9 \cite{rosenthal-etal-2014-semeval}, SemEval 2015 Task 10 \cite{rosenthal-etal-2015-semeval}, and SemEval 2018 Task 1 \cite{mohammad-etal-2018-semeval}. (The last of which also includes and evaluation component for demographic bias in sentiment analysis systems.)
See \citet{rottger2020hatecheck} for work on creating separate diagnostic datasets for various types of hate speech.
See Google’s recommendations on best practices on metrics and testing (https://ai.google/responsibilities/responsible-ai-practices)\\ 

\noindent \textbf{H. Beyond Metrics}\\[2pt]
\noindent \textit{Are we even measuring the right things?}\\[-3pt]

\noindent \textbf{\#36. Interpretability, Explainability:} As ML systems are deployed more widely and impact a greater sphere of our lives, there is a growing understanding that these systems can be flawed to varying degrees. One line of approach in understanding and addressing these flaws is to develop interpretable or explainable models. Interpretability and explainability each have been defined in a few different ways in the literature, but at the heart of the definitions is the idea that we should be able to understand why a system is making a certain prediction: what pieces of evidence are contributing to the decision and to what degree? That way, humans can better judge how valid a particular prediction is, better judge how accurate the model is for certain kinds of input, and even how accurate the system is in general and over time.

In line with this, AER systems should have components that depict why they are making certain predictions for various inputs. 
As described in the \citet{luo2021local} survey,
such components can be viewed from several perspectives, including:\\[-20pt]
\begin{itemize}
    \item are the explanations meant for the scientist/engineer or to a lay person?
    \vspace*{-1mm}
    \item are the explanations faithful (accurate reflections of system behavior)?
    \vspace*{-1mm}
    \item are the explanations easily comprehensible?
    \vspace*{-1mm}
    \item to what extent do people trust the explanations?
\end{itemize}
\vspace*{-3mm}
\noindent Responsible research and product development entails actively considering various explainability strategies at the very outset of the project. This includes, where appropriate, specifically choosing an ML model that lends itself to better interpretability, running ablation and disaggregation experiments, running data perturbation and adversarial testing experiments, and so on.\\


\noindent \textbf{\#37. Visualization:} Visualizations help convey trends in emotions and sentiments, and are common in the emotion analysis of streams of data such as tweet streams, novels, newspaper headlines, etc. There are several considerations when developing visualizations that impact the extent to which they are effective, convey key trends, and the extent to which they may be misleading:\\[-20pt]
\begin{itemize}
    \item It is almost always important to not only show the broad trends but also to allow the user to drill down to the source data that is driving the trend.
    \vspace*{-1mm}
    \item Summarize the data driving the trend, for example through treemaps of the most frequent emotion words and phrases in the data.
    \vspace*{-1mm}
    \item Interactive visualizations allow users to explore different trends in the data and even drill down to the source data that is driving the trends.
\end{itemize}
\vspace*{-3mm}
\noindent See work on visualizing emotions and sentiment 
\cite{mohammad-2011-upon,dwibhasi2015analyzing,kucher2018visual,fraser-etal-2019-feel,gallagher2021generalized}.\\

\noindent \textbf{\#38. Safeguards and guard rails:} Devote time and resources to identify how the system can be misused and how the system may cause harm because of it’s inherent biases and limitations. Identify steps that can be taken to mitigate these harms.\\

\noindent \textbf{\#39. Recognize that there will be harms even when the system works ``correctly”:} Provide a mechanism for users to report issues. Have resources in place to deal with unanticipated harms. Document societal impacts, including both benefits and harms.\\

\noindent \textbf{\#40. Contestability and Recourse:} \citet{mulligan2019shaping} argue that contestability---the mechanisms made available to challenge the predictions of an AI system---are more important and beneficial than transparency/explainability. Not only do they allow people to challenge the decisions made by a system, they also invite participation in the understanding of how machine learning systems work and their limitations. See Google’s The What-If Tool as an example of how people are invited to explore ML systems by changing inputs (without needing to do any coding) \cite{what_if_2018}.
AER systems are encouraged to produce similar tools, for example:\\[-20pt]
\begin{itemize}
    \item tools that allow one to see counterfactuals—given a data point, what is the closest other data point for which the system predicts a different label; tools that allow one to try out various input conditions/features to see what help obtain the desired classification label. 
\vspace*{-1mm}
    \item tools that allow one to see classification accuracies on different demographics and the impact of different classifier parameters and thresholds on these scores.
\vspace*{-1mm}
    \item tools that allow one to see confidence of the classifier for a given prediction and the features that were primarily responsible for the decision.
\end{itemize}
\vspace*{-3mm}
\noindent See \citet{denton2020bringing} for ideas on on participatory dataset creation and management.\\

\noindent \textbf{\#41. Be wary of Ethics Washing:} As we push farther into incorporating ethical practices in our projects, we need to be wary of inauthentic and cursory attention to ethics for the sake of appearances. This VentureBeat article \cite{johnson_2019} presents some nice tips to avoid ethics washing, including: “Welcome ‘constructive dissent’ and uncomfortable conversations”, “Don’t ask for permission to get started”, “Share your shortcomings”, “Be prepared for gray area decision-making”, and “Ethics has few clear metrics”.


\subsubsection{IMPLICATIONS FOR PRIVACY, SOCIAL GROUPS} (Nine considerations.)\\
\label{sec:privacysg}


\noindent \textbf{I. Implications for Privacy}\\
\noindent \textit{(Cuts across Task Design, Data, Method, Impact and Evaluation)}\\

\noindent \textbf{\#42. Privacy and Personal Control:} As noted privacy expert, 
Dr. Ann Cavoukian, puts it: privacy is not about hiding information or secrecy. It is about choice, ``You have to be the one to make the decision."
    Individuals may not want their emotions to be inferred. 
Applying emotion detection systems en masse---gathering emotion information continuously, without meaningful consent, is an invasion of privacy, harmful to the individual, and dangerous to society. (See report  created for the members of the European Parliament \cite{woensel_nevil_2019}).
Follow the seven principles of privacy by design \cite{schaar2010privacy}:
    Proactive not Reactive (preventative not remedial), 
    Privacy as the Default,
    Privacy Embedded into Design,
    Full Functionality (positive-sum, not zero-sum),
    End-to-End Security (full lifecycle),
    Visibility and Transparency, and
    Respect for User Privacy (keep it user-centric).
See also privacy-preserving work on sentiment by \citet{thaine-penn-2021-chinese}.\\

\noindent \textbf{\#43. Group Privacy and Soft Biometrics:} \citet{floridi2014open} argues that many of our conversations around privacy are far too focused on individual privacy and ignore group privacy — the rights and protections we need as a group.\\[-20pt]
\begin{quote}
    \textit{There are very few Moby-Dicks. Most of us are sardines. The individual sardine may believe that the encircling net is trying to catch it. It is not. It is trying to catch the whole shoal. It is therefore the shoal that needs to be protected, if the sardine is to be saved.} — \citet{floridi2014open}
\end{quote}
\vspace*{-3mm}
\noindent The idea of group privacy becomes especially important in the context of soft-biometrics such as traits and preferences determined through AER that are not intended to be able to identify individuals, but rather identify groups of people with similar characteristics. See \citet{mcstay2020emotional} for further discussions on the implications of AER on group privacy and how companies are using AER to determine group preferences, even though a large number of people disfavour such profiling.\\

\noindent \textbf{\#44. Mass Surveillance versus Right to Privacy, Right to Freedom of Expression, and Right to Protest:} Emotion recognition, sentiment analysis, and stance detection can be used for mass surveillance by companies and governments (often without meaningful consent). There is low awareness in people that their information (e.g., what they say or click on an online platform) can be used against their best interest. Often people do not have meaningful choices regarding privacy when they use online platforms.
In extreme cases, as in the case of authoritarian governments, this can lead to dramatic curtailing of freedoms of expression and the right to protest \cite{article19_2021,wakefield_2021}.\\

\noindent \textbf{\#45. Right Against Self-Incrimination:} In a number of countries around the world, the accused are given legal rights against self-incrimination. However, automatic methods of emotion, stance, and deception detection can potentially be used to circumvent such protections. (See \citet{article19_2021} page 37.)\\

\noindent \textbf{\#46. Right to Non-Discrimination:} 
Automatic methods of emotion, stance, and deception detection can sometimes systematically discriminate based on these protected categories such as race, gender, and religion. Even if ML systems are not fed race or gender information directly, studies have shown that they often pick up on proxy attributes for these categories. Report disaggregated results as appropriate.\\[2pt]

\noindent \textbf{J. Implications for Social Groups}\\[2pt]
\noindent \textit{(Cuts across Task Design, Data, Method, Impact and Evaluation)}\\


\noindent \textbf{\#47. Disaggregation:} Society has often viewed different groups differently (because of their race, gender, income, language, etc.), imposing unequal social and power structures \cite{lindsey2015sociology}. Even when the biases are not conscious, the unique needs of different groups is often overlooked. For example, \citet{perez2019invisible} discusses, through numerous examples, how there is a considerable lack of disaggregated data for women and how that is directly leading to negative outcomes in all spheres of their lives, including health, income, safety, and the degree to which they succeed in their endeavors. This holds true (perhaps even more) for transgender people. Thus emotion researchers should consider the value of disaggregation at various levels, including:\\[-20pt]
\begin{itemize}
    \item When creating datasets: Obtain annotations from a diverse group of people. Report aggregate-level demographic information. Rather than only labeling instances with the majority vote, consider the value of providing multiple sets of labels as per each of the relevant and key demographic groups.
\vspace*{-1mm}
    \item When testing hypotheses or drawing inferences about language use: Consider also testing the hypotheses disaggregated for each of the relevant and key demographic groups.
\vspace*{-1mm}
    \item When building automatic prediction systems: Report performance disaggregated for each of the relevant and key demographic groups. (See work on model cards \citet{mitchell2019model}. See how sentiment analysis systems can be systematically biased \cite{kiritchenko-mohammad-2018-examining}.)\\
\end{itemize}
\vspace*{-3mm}
\noindent \textbf{\#48. Intersectional Invisibility in Research:} Intersectionality refers to the complex ways in which different group identities such as race, class, neurodiversity, and gender overlap to amplify discrimination or disadvantage. \citet{purdie2008intersectional} argue how people with multiple group identities are often not seen as prototypical members of any of their groups and thus are subject to, what they, call intersectional invisibility---omissions of their experiences in historical narratives and cultural representation, lack of support from advocacy groups, and mismatch with existing anti-discrimination frameworks. Many of the forces that lead to such invisibility (e.g., not being seen as prototypical members of a group) along with other notions common in the quantitative research paradigm (e.g., the predilection to work on neat, non-overlapping, populous categories) lead to intersectional invisibility in research. As ML/NLP researchers, we should be cognizant of such blind spots 
and work to address these gaps. Further, new ways of doing research that address the unique challenges of doing intersectional research need to be valued and encouraged.\\[-2pt]

\noindent \textbf{\#49. Reification and Essentialization:} Some demographic variables are essentially, or in big part, social constructs. Thus, work on disaggregation can sometimes reinforce false beliefs that there are innate differences across different groups or that some features are central for one to belong to a social category. Thus it is imperative to contextualize work on disaggregation. For example, by impressing on the reader that even though race is a social construct, the impact of people’s perceptions and behavior around race lead to very real-world consequences.\\[-2pt]

\noindent \textbf{\#50. Attributing People to Social Groups:} In order to be able to obtain disaggregated results, sometimes one  needs access to demographic information. 
This of course leads to considerations such as: whether they are providing meaningful consent to the collection of such data and whether the data being collected in a manner that respects their privacy, their autonomy (e.g., can they choose to delete their information later), and dignity (e.g., allowing self-descriptions).
Challenges persist in terms of how to design effective and inclusive questionnaires \cite{bauer2017transgender}.
Further, even with self-report textboxes that give the respondent the primacy and autonomy to express their race, gender, etc.\@, downstream research often ignores such data or combines information in ways beyond the control of the respondent \cite{keyes_2019}.

Some work tries to infer aggregate-level group statistics automatically. For example, inferring race, gender, etc.\@ from cues such as the type of language used, historical name-gender associations, etc. to do disaggregated analysis. However, such approaches are fraught with ethical concerns such as misgendering, essentialization, and reification. Further, historically, people have been marginalized because of their social category, and so methods that try to detect these categories raise legitimate and serious concerns of abuse, erasure, and perpetuating stereotypes.

In many cases, it may be more appropriate to perform disaggregated analysis on something other than a social category. For example, when testing face recognition systems, it might be more appropriate to test the system performance on different skin tones (as opposed to race). Similarly, when working on language data, it might be more appropriate to analyze data partitioned by linguistic gender (as opposed to social gender). See \citet{cao2021toward} for a useful discussion on linguistic vs. social gender and also for a great example to create more inclusive data for research.\\


\section{In Summary}
This paper aggregates and organizes various ethical considerations relevant to automatic emotion recognition, drawn from the wider AI Ethics and Affective Computing literature. It includes brief sections on the modalities of information, task, and applications of AER to set the context. Then it presents fifty ethical considerations grouped thematically. 
Notably, the sheet fleshes out assumptions hidden in how AER is commonly framed, and in the choices often made regarding the data, method, and evaluation. Special attention is paid to the implications of AER  on privacy and social groups.
It discusses how these considerations manifest within AER and outlines best practices for responsible research.
A succinct list of key recommendations for responsible AER discussed in the paper is provided in the Appendix.

The objective of the sheet is to encourage practitioners to think in more detail and at the very outset: why to automate, how to automate, and how to judge success based on broad societal implications. 
I hope that it will help engage the various stakeholders of AER with each other; help stakeholders challenge assumptions made by researchers and developers; and help develop appropriate harm mitigation strategies.
Additionally, for those that are new to emotion recognition, the ethics sheet acts as a useful introductory document  (complementing survey articles).

As an expert on a technology, an often overlooked and undervalued responsibility is to convey its broad societal impacts 
to those that deploy the technology, those that make policy decisions about the technology, and the society at large. 
I hope that this sheet helps to that end for emotion recognition, and also spurs the wider community to ask and document: \textit{What ethical considerations apply to my task?} 


\begin{acknowledgments}
 I am grateful to Annika Schoene, Mallory Feldman, and Tara Small for their belief and encouragement in the early days of this project.
 Many thanks to Mallory Feldman (Carolina Affective Neuroscience Lab, UNC) for discussions on the psychology and complexity of emotions. 
  Many thanks to Annika Schoene, Mallory Feldman, Roman Klinger, Rada Mihalcea, Peter Turney, Barbara Plank, Malvina Nissim, Viviana Patti, Maria Liakata, and Emily Mower Provost for discussions about ethical considerations for emotion recognition and thoughtful comments. Many thanks to Tara Small, Emily Bender, Esma Balkir, Isar Nejadgholi, Patricia Thaine, Brendan O’Connor, Cyril Goutte, Eric Joanis, Joel Martin, Roland Kuhn, and Sowmya Vajjala for thoughtful comments on the blog post on this work.
 \end{acknowledgments}

\vspace*{10mm}
\noindent {\bf APPENDIX: Recommendations for Responsible AER}\\

\noindent Below is a list of key recommendations for responsible AER discussed earlier in the context of various ethical considerations. They are compiled here for easy access. Note that adhering to these recommendations does not guarantee ``ethicalness''; nor do these recommendations apply to all contexts. They are guidelines meant to help responsible development and use of AER systems. Particular development or deployment contexts entail further considerations and steps to address them.

{\small
\vspace*{5mm}
\noindent \textit{Task Design}
\begin{enumerate}

\item Center the people, especially marginalized and disadvantaged communities, such that they are not mere passive subjects but rather have the agency to shape the design process. 

\item Ask who will benefit from this work and who will not? Will this work shift power from those who already have a lot of power to those that have less power? How can the task be designed so that it helps those that are most in need ?

\item Ask how the AER design will impact people in the context of  neurodiversity, alexithymia, and autism spectrum. 

\item Carefully consider what emotion task should be the focus of the work (whether conducting a human-annotation study or building an automatic prediction model). Different emotion tasks entail different ethical considerations.
Communicate the nuance of exactly what emotions are being captured to the stakeholders. Not doing so will mean will lead to the misuse and misinterpretation of one’s work.

\item AER systems should not claim to determine one's emotional state from their utterance, facial expression, gait, etc. At best, AER systems capture what one is trying to convey or what is perceived by the listener/viewer, and even there, given the complexity of human expression, they are often inaccurate.

\item Even when AER systems attempt to determine the emotional state of a person (or a group) over time (drawing inferences at aggregate level from large amounts of data), such as studies on public health listed in \S3.3, it is best to be cautious when making claims. Use AER as one source of evidence amongst many (and involve relevant expertise; e.g., from public health and psychology).

\item Lay out the theoretical foundations for the task from relevant research fields such as psychology, linguistics, and sociology, and relate the opinions of relevant domain experts to the task formulation. Realize that it is impossible to capture the full emotional experience of a person.
\item Do not refer to some emotions as basic emotions, unless you mean to convey your belief in the Basic Emotions Theory. Careless endorsement of theories can lead to the perpetuation of belief in ideas that are actively harmful (such as suggesting we can determine internal state from outward appearance — physiognomy).
\item Realize that various ethical concerns, including privacy, manipulation, bias, and free speech, are further exacerbated when systems that act on individuals. Take steps such as anonymization and realizing information at aggregate levels.

\item Think about how the AER system can be misused, and how that can be minimized.
\item Use AER as one source of information among many.
\item Do not use AER for fully automated decision making. AER may be used to assist humans in making decisions, coming up with ideas, suggesting where to delve deeper, and sparking their imagination. Consider also the risk of the system inappropriately biasing the human decision makers.
\item Disclose to all stakeholders the decisions that are being made (in part or wholly) by automation. Provide mechanisms for the user to understand why relevant predictions were made, and also to contest the decisions.
\end{enumerate}

\noindent \textit{Data}
\begin{enumerate}
  \setcounter{enumi}{13}
\item Examine the choice of data used by AER systems across various dimensions: size of data; whether it is custom data or data obtained from an online platform; less private/sensitive data or more private/sensitive data; what languages are represented; degree of documentation; and so on. 
\item Expressions of emotions through language are highly variable: Different people express the same emotion differently; the same text may convey different emotions to different people. This variability should also be taken into consideration when building datasets, systems, and choosing where to deploy the systems. 
\item Variability is common not just for emotions but also for natural language. People convey meaning in many different ways. There is usually no one “correct” way of articulating our thoughts.
\item Aim to obtain useful level of emotion recognition capabilities without having systematic gaps that convey a strong sense of emotion-expression normativeness.
\item When using language models or emotion datasets, avoid perpetuating stereotypes of how one group of people perceive another group.
\item Obtain data from a diverse set of sources. Report details of the sources.
\item When creating emotion datasets, limit the number of instances included per person. Mohammad and Kiritchenko (2018) kept one tweet for every query term and tweeter combination when studying relationships between affect categories (data also used in a shared task on emotions). Kiritchenko et al., (2020) kept at most three tweets per tweeter when studying expressions of loneliness.
\item Obtain annotations from a diverse set of people. Report aggregate-level demographic information of the annotators.
\item In emotion and language data, often there are no “correct” answers. Instead, it is a case of some answers being more appropriate than others. And there can be multiple appropriate answers.
\item Part of conveying that there is no one “correct” answer is to convey how the dataset is situated in many parameters, including: who annotated it, the precise annotation instructions, what data was presented to the annotators (and in what form), and when the data was annotated.
\item Release raw data annotations as well as any aggregations of annotations.
\item If using majority voting, acknowledge its limitations. 
\item Explore statistical approaches to finding multiple appropriate labels. 
\item Employ manual and automatic checks to determine whether the human annotations have also captures inappropriate biases. 
Such biases may be useful for some projects (e.g., work studying such biases), but not for others. Warn users appropriately and deploy measures to mitigate their impact.
\item When using any dataset, devote time and resources to study who is included in the dataset and whose voices are missing. Take corrective action as appropriate.
\item Keep a portion of your funding for work with marginalized communities and for work on less-researched languages.
\item Systems that are to be deployed to handle open-domain data should be trained on many diverse datasets and tested on many datasets that are quite different from the training datasets.
\item Ensure that the terms of service of the source platforms are not violated: e.g., data scraping is allowed and data redistribution is allowed (in raw form or through ids). Check the platform terms of service. Ensure compliance with the robot exclusion protocol.
Take actions to anonymize data when dealing with sensitive or private data; e.g., scrub identifying information. Choose to not work with a dataset if adequate safeguards cannot be placed.
\item Proposals of data annotation efforts that may impact the well-being of annotators should first be submitted for approval to one’s Research Ethics Board (REB) / Institutional Research Board (IRB). The board will evaluate and provide suggestions so that the work complies with the required ethics standards.
\item An excellent jumping off point for further information on ethical conduct of research involving human subjects is The Belmont Report. The guiding principles they proposed are Respect for Persons, Beneficence, and Justice.
\end{enumerate}
\noindent \textit{Method}
\begin{enumerate}
  \setcounter{enumi}{33}
\item Examine choice of method across various dimensions such as interpretability, privacy concerns, energy efficiency, data needs, etc . Focusing on fewer dimensions may be okay in a research system, but widely deployed systems often require a good balance across the many dimensions.
AI methods tend to work well for people that are well-represented in the data (raw and annotated), but not so well for others. Documenting who is left-out is valuable. Explore alternative methods that are more inclusive. 
Consider how the data collection and machine learning setups can be addressed to avoid spurious correlations, especially correlations that perpetuate racism, sexism, and stereotypes.
\item Systems are often trained on static data from the past. However, perceptions, emotions, and behavior change with time. Consider how automatic systems may make inappropriate predictions on current data.
\item Consider the system deployment context to determine what levels of emotional management and meaningful consent are appropriate.
\item Consider the carbon footprint of your method and value efficiency as a contribution. Report costs per example, size of training set, number of hyperparameters, and budget-accuracy curves.
\end{enumerate}
\noindent \textit{Impact and Evaluation}
\begin{enumerate}
  \setcounter{enumi}{37}
\item Consider whether the chosen metrics are measuring what matters.
\item Some methods can be unreliable or systematically inaccurate for certain groups of people, races, genders, people with health conditions, people from different countries, etc. Determine and present disaggregated accuracies.
Test the system on many different datasets that explore various input characteristics. 
\item Responsible research and product development entails actively considering various explainability strategies at the very outset of the project. This includes, where appropriate, specifically choosing an ML model that lends itself to better interpretability, running ablation and disaggregation experiments, running data perturbation and adversarial testing experiments, and so on.
\item When visualizing emotions, it is almost always important to not only show the broad trends but also to allow the user to drill down to the source data that is driving the trend. One can also summarize the data driving the trend, for example through treemaps of the most frequent emotion words.
\item Devote time and resources to identify how the system can be misused and how the system may cause harm because of it’s inherent biases and limitations. Recognize that there will be harms even when the system works “correctly”.
Identify steps that can be taken to mitigate these harms.
\item Provide mechanisms for contestability that not only allow people to challenge the decisions made by a system about them, but also invites participation in the understanding of how machine learning systems work and it limitations.
\end{enumerate}
\noindent \textit{Implications for Privacy}
\begin{enumerate}
  \setcounter{enumi}{43}
\item Privacy is not about secrecy. It is about personal choice. Follow Dr. Cavoukian’s seven principles of privacy by design.
\item Consider that people might not want their emotions to be inferred. Applying emotion detection systems en masse --- gathering emotion information continuously, without meaningful consent, is an invasion of privacy, harmful to the individual, and dangerous to society.
\item Soft-biometrics also have privacy concerns. Consider implications of AER on group privacy and that a large number of people disfavour such profiling.
\item Obtain meaningful consent as appropriate for the context. Working with more sensitive and more private data requires a more involved consent process where the user understands the privacy concerns and willingly provides consent.
Consider harm mitigation strategies such as: anonymization techniques and differential privacy. Beware that these can vary in effectiveness.
\item Plan for how to keep people's information secure.
\item Obtain permission for secondary use or if you intend to distribute the data.
\item When working out the privacy–benefit tradeoffs, consider who will really benefit from the technology. Especially consider whether those who benefit are people with power or those with less power. Also, as Dr. Cavoukian says, often privacy and benefits can both be had, “it is not a zero-sum game”.
\item Consider implications of AER for mass surveillance and how that undermines right to privacy, right to freedom of expression, right to protest, right against self-incrimination, and right to non-discrimination.
\end{enumerate}
\noindent \textit{Implications for Social Groups}
\begin{enumerate}
  \setcounter{enumi}{51}
\item When creating datasets, obtain annotations from a diverse group of people. Report aggregate-level demographic information. Rather than only labeling instances with the majority vote, consider the value of providing multiple sets of labels as per each of the relevant and key demographic groups.
\item When testing hypotheses or drawing inferences about language use, consider also testing the hypotheses disaggregated for each of the relevant demographic groups.
\item When building automatic prediction systems, evaluate and report performance disaggregated for each of the relevant demographic groups.
\item Consider and report the implication of the AER system on intersectionality.
\item Contextualize work on disaggregation: for example, by impressing on the reader that even though race is a social construct, the impact of people’s perceptions and behavior around race lead to very real-world consequences.
\item Obtaining demographic information requires careful and thoughtful considerations such as: whether people are providing meaningful consent to the collection of such data and whether the data being collected in a manner that respects their privacy, their autonomy (e.g., can they choose to delete their information later), and dignity (e.g., allowing self-descriptions).
\end{enumerate}
}

\starttwocolumn
\bibliography{compling_style}

\begin{thebibliography}{131}
\expandafter\ifx\csname natexlab\endcsname\relax\def\natexlab#1{#1}\fi

\bibitem[{Agrawal, Batra, and Parikh(2016)}]{agrawal2016analyzing}
Agrawal, Aishwarya, Dhruv Batra, and Devi Parikh. 2016.
\newblock Analyzing the behavior of visual question answering models.
\newblock \emph{arXiv preprint arXiv:1606.07356}.

\bibitem[{AI2(2019)}]{ai2_2019}
AI2. 2019.
\newblock Crowdsourcing: Pricing ethics and best practices.
\newblock Medium.
  \url{https://medium.com/ai2-blog/crowdsourcing-pricing-ethics-and-best-practices-8487fd5c9872}.

\bibitem[{Arcas, Mitchell, and Todorov(2017)}]{physiognomy_2017}
Arcas, Blaise, Margaret Mitchell, and Alexander Todorov. 2017.
\newblock Physiognomy's new clothes.
\newblock Medium.
  \url{https://medium.com/@blaisea/physiognomys-new-clothes-f2d4b59fdd6a}.

\bibitem[{Aroyo and Welty(2015)}]{aroyo2015truth}
Aroyo, Lora and Chris Welty. 2015.
\newblock Truth is a lie: Crowd truth and the seven myths of human annotation.
\newblock \emph{AI Magazine}, 36(1):15--24.

\bibitem[{ARTICLE19(2021)}]{article19_2021}
ARTICLE19. 2021.
\newblock Emotional entanglement: China’s emotion recognition market and its
  implications for human rights.
\newblock
  \url{https://www.article19.org/wp-content/uploads/2021/01/ER-Tech-China-Report.pdf}.

\bibitem[{Auxier and Anderson(2021)}]{auxier2021social}
Auxier, Brooke and Monica Anderson. 2021.
\newblock Social media use in 2021.
\newblock \emph{Pew Research Center}.

\bibitem[{Bagby, Parker, and Taylor(1994)}]{bagby1994twenty}
Bagby, R~Michael, James~DA Parker, and Graeme~J Taylor. 1994.
\newblock The twenty-item {T}oronto {A}lexithymia scale: I. item selection and
  cross-validation of the factor structure.
\newblock \emph{Journal of psychosomatic research}, 38(1):23--32.

\bibitem[{Bamberg(1997)}]{bamberg1997language}
Bamberg, Michael. 1997.
\newblock Language, concepts and emotions: The role of language in the
  construction of emotions.
\newblock \emph{Language sciences}, 19(4):309--340.

\bibitem[{Barrett(2017{\natexlab{a}})}]{barrett2017emotions}
Barrett, Lisa~Feldman. 2017{\natexlab{a}}.
\newblock \emph{How emotions are made: The secret life of the brain}.
\newblock Houghton Mifflin Harcourt.

\bibitem[{Barrett(2017{\natexlab{b}})}]{barrett2017theory}
Barrett, Lisa~Feldman. 2017{\natexlab{b}}.
\newblock The theory of constructed emotion: an active inference account of
  interoception and categorization.
\newblock \emph{Social cognitive and affective neuroscience}, 12(1):1--23.

\bibitem[{Barrett et~al.(2019)Barrett, Adolphs, Marsella, Martinez, and
  Pollak}]{barrett2019emotional}
Barrett, Lisa~Feldman, Ralph Adolphs, Stacy Marsella, Aleix~M Martinez, and
  Seth~D Pollak. 2019.
\newblock Emotional expressions reconsidered: Challenges to inferring emotion
  from human facial movements.
\newblock \emph{Psychological science in the public interest}, 20(1):1--68.

\bibitem[{Basile(2020)}]{basile2020s}
Basile, Valerio. 2020.
\newblock It’s the end of the gold standard as we know it. on the impact of
  pre-aggregation on the evaluation of highly subjective tasks.
\newblock In \emph{2020 AIxIA Discussion Papers Workshop, AIxIA 2020 DP},
  volume 2776, pages 31--40, CEUR-WS.

\bibitem[{Bauer et~al.(2017)Bauer, Braimoh, Scheim, and
  Dharma}]{bauer2017transgender}
Bauer, Greta~R, Jessica Braimoh, Ayden~I Scheim, and Christoffer Dharma. 2017.
\newblock Transgender-inclusive measures of sex/gender for population surveys:
  Mixed-methods evaluation and recommendations.
\newblock \emph{PloS one}, 12(5):e0178043.

\bibitem[{Bender et~al.(2021)Bender, Gebru, McMillan-Major, and
  Shmitchell}]{bender2021dangers}
Bender, Emily~M, Timnit Gebru, Angelina McMillan-Major, and Shmargaret
  Shmitchell. 2021.
\newblock On the dangers of stochastic parrots: Can language models be too big?
\newblock In \emph{Proceedings of the 2021 ACM Conference on Fairness,
  Accountability, and Transparency}, pages 610--623.

\bibitem[{Bender and Koller(2020)}]{bender-koller-2020-climbing}
Bender, Emily~M. and Alexander Koller. 2020.
\newblock Climbing towards {NLU}: {On} meaning, form, and understanding in the
  age of data.
\newblock In \emph{Proceedings of the 58th Annual Meeting of the Association
  for Computational Linguistics}, pages 5185--5198, Association for
  Computational Linguistics, Online.

\bibitem[{Bertilsdotter~Rosqvist et~al.(2019)Bertilsdotter~Rosqvist, Kourti,
  Jackson-Perry, Brownlow, Fletcher, Bendelman, and
  O'Dell}]{bertilsdotter2019doing}
Bertilsdotter~Rosqvist, Hanna, Marianthi Kourti, David Jackson-Perry, Charlotte
  Brownlow, Kirsty Fletcher, Daniel Bendelman, and Lindsay O'Dell. 2019.
\newblock Doing it differently: Emancipatory autism studies within a
  neurodiverse academic space.
\newblock \emph{Disability \& Society}, 34(7-8):1082--1101.

\bibitem[{Bisk et~al.(2020)Bisk, Holtzman, Thomason, Andreas, Bengio, Chai,
  Lapata, Lazaridou, May, Nisnevich, Pinto, and
  Turian}]{bisk-etal-2020-experience}
Bisk, Yonatan, Ari Holtzman, Jesse Thomason, Jacob Andreas, Yoshua Bengio,
  Joyce Chai, Mirella Lapata, Angeliki Lazaridou, Jonathan May, Aleksandr
  Nisnevich, Nicolas Pinto, and Joseph Turian. 2020.
\newblock Experience grounds language.
\newblock In \emph{Proceedings of the 2020 Conference on Empirical Methods in
  Natural Language Processing (EMNLP)}, pages 8718--8735, Association for
  Computational Linguistics, Online.

\bibitem[{Bissoto, Valle, and Avila(2020)}]{bissoto2020debiasing}
Bissoto, Alceu, Eduardo Valle, and Sandra Avila. 2020.
\newblock Debiasing skin lesion datasets and models? not so fast.
\newblock In \emph{Proceedings of the IEEE/CVF Conference on Computer Vision
  and Pattern Recognition Workshops}, pages 740--741.

\bibitem[{Born et~al.(2021)Born, Morris, Diaz, and Anderson}]{Born21}
Born, Georgina, Jeremy Morris, Fernando Diaz, and Ashton Anderson. 2021.
\newblock Artificial {I}ntelligence, music recommensation, and the curation of
  culture.
\newblock Technical report.

\bibitem[{Boyle(1995)}]{Boyle95}
Boyle, Gregory~J. 1995.
\newblock Myers--{B}riggs type indicator ({MBTI}): Some psychometric
  limitations.
\newblock \emph{Humanities \& Social Sciences papers}, 30.

\bibitem[{Brosnan et~al.(2017)Brosnan, Holt, Yuill, Good, and
  Parsons}]{brosnan2017beyond}
Brosnan, Mark, Samantha Holt, Nicola Yuill, Judith Good, and Sarah Parsons.
  2017.
\newblock Beyond autism and technology: Lessons from neurodiverse populations.
\newblock \emph{Journal of Enabling Technologies}.

\bibitem[{Buolamwini and Gebru(2018)}]{buolamwini2018gender}
Buolamwini, Joy and Timnit Gebru. 2018.
\newblock Gender shades: Intersectional accuracy disparities in commercial
  gender classification.
\newblock In \emph{Conference on fairness, accountability and transparency},
  pages 77--91, PMLR.

\bibitem[{Cao and Daum{\'e}(2021)}]{cao2021toward}
Cao, Yang~Trista and Hal Daum{\'e}. 2021.
\newblock Toward gender-inclusive coreference resolution: An analysis of gender
  and bias throughout the machine learning lifecyle.
\newblock \emph{Computational Linguistics}, pages 1--47.

\bibitem[{Chancellor et~al.(2019)Chancellor, Birnbaum, Caine, Silenzio, and
  De~Choudhury}]{10.1145/3287560.3287587}
Chancellor, Stevie, Michael~L. Birnbaum, Eric~D. Caine, Vincent M.~B. Silenzio,
  and Munmun De~Choudhury. 2019.
\newblock A taxonomy of ethical tensions in inferring mental health states from
  social media.
\newblock In \emph{Proceedings of the Conference on Fairness, Accountability,
  and Transparency}, FAT* '19, page 79–88, Association for Computing
  Machinery, New York, NY, USA.

\bibitem[{Checco et~al.(2017)Checco, Roitero, Maddalena, Mizzaro, and
  Demartini}]{checco2017let}
Checco, Alessandro, Kevin Roitero, Eddy Maddalena, Stefano Mizzaro, and
  Gianluca Demartini. 2017.
\newblock Let's agree to disagree: Fixing agreement measures for crowdsourcing.
\newblock In \emph{Proceedings of the Fifth AAAI Conference on Human
  Computation and Crowdsourcing}, pages 11--20.

\bibitem[{Chomsky(1975)}]{chomsky1975reflections}
Chomsky, Noam. 1975.
\newblock \emph{Reflections on language}.
\newblock Pantheon.

\bibitem[{Chomsky(2014)}]{chomsky2014aspects}
Chomsky, Noam. 2014.
\newblock \emph{Aspects of the Theory of Syntax}, volume~11.
\newblock MIT press.

\bibitem[{Cobb-Clark and Schurer(2012)}]{COBBCLARK201211}
Cobb-Clark, Deborah~A. and Stefanie Schurer. 2012.
\newblock The stability of big-five personality traits.
\newblock \emph{Economics Letters}, 115(1):11--15.

\bibitem[{De~Choudhury et~al.(2013)De~Choudhury, Gamon, Counts, and
  Horvitz}]{de2013predicting}
De~Choudhury, Munmun, Michael Gamon, Scott Counts, and Eric Horvitz. 2013.
\newblock Predicting depression via social media.
\newblock In \emph{Seventh international AAAI conference on weblogs and social
  media}, pages 128--137.

\bibitem[{De~Cicco, Palumbo et~al.(2020)}]{de2020should}
De~Cicco, Roberta, Riccardo Palumbo, et~al. 2020.
\newblock Should a chatbot disclose itself? {I}mplications for an online
  conversational retailer.
\newblock In \emph{International Workshop on Chatbot Research and Design},
  pages 3--15, Springer.

\bibitem[{Denton et~al.(2020)Denton, Hanna, Amironesei, Smart, Nicole, and
  Scheuerman}]{denton2020bringing}
Denton, Emily, Alex Hanna, Razvan Amironesei, Andrew Smart, Hilary Nicole, and
  Morgan~Klaus Scheuerman. 2020.
\newblock Bringing the people back in: Contesting benchmark machine learning
  datasets.
\newblock \emph{arXiv preprint arXiv:2007.07399}.

\bibitem[{Dickson(2018)}]{dickson_2018}
Dickson, Ben. 2018.
\newblock Why {AI} must disclose that it's {AI}.
\newblock PC Magazine.
  \url{https://www.pcmag.com/opinions/why-ai-must-disclose-that-its-ai}.

\bibitem[{Dolmaya(2011)}]{dolmaya2011ethics}
Dolmaya, Julie~McDonough. 2011.
\newblock The ethics of crowdsourcing.
\newblock \emph{Linguistica Antverpiensia, New Series--Themes in Translation
  Studies}, (10).

\bibitem[{Dwibhasi et~al.(2015)Dwibhasi, Jami, Lanka, and
  Chakraborty}]{dwibhasi2015analyzing}
Dwibhasi, Sharat, Dheeraj Jami, Shivkanth Lanka, and Goutam Chakraborty. 2015.
\newblock Analyzing and visualizing the sentiments of {E}bola outbreak via
  tweets.
\newblock In \emph{Proceedings of the SAS Global Forum, Dallas, TX, USA}, pages
  26--29.

\bibitem[{Eichstaedt et~al.(2015)Eichstaedt, Schwartz, Kern, Park, Labarthe,
  Merchant, Jha, Agrawal, Dziurzynski, Sap
  et~al.}]{eichstaedt2015psychological}
Eichstaedt, Johannes~C, Hansen~Andrew Schwartz, Margaret~L Kern, Gregory Park,
  Darwin~R Labarthe, Raina~M Merchant, Sneha Jha, Megha Agrawal, Lukasz~A
  Dziurzynski, Maarten Sap, et~al. 2015.
\newblock Psychological language on {T}witter predicts county-level heart
  disease mortality.
\newblock \emph{Psychological science}, 26(2):159--169.

\bibitem[{Ekman(1992)}]{ekman1992there}
Ekman, Paul. 1992.
\newblock Are there basic emotions?
\newblock \emph{Psychological Review}, 99(3):550--553.

\bibitem[{Ekman and Davidson(1994)}]{ekman1994nature}
Ekman, Paul~Ed and Richard~J Davidson. 1994.
\newblock \emph{The nature of emotion: Fundamental questions.}
\newblock Oxford University Press.

\bibitem[{Ervin-Tripp(1973)}]{ervin1973some}
Ervin-Tripp, Susan. 1973.
\newblock Some strategies for the first two years.
\newblock In \emph{Cognitive Development and Acquisition of Language}.
  Elsevier, pages 261--286.

\bibitem[{Fletcher-Watson et~al.(2019)Fletcher-Watson, Adams, Brook, Charman,
  Crane, Cusack, Leekam, Milton, Parr, and Pellicano}]{fletcher2019making}
Fletcher-Watson, Sue, Jon Adams, Kabie Brook, Tony Charman, Laura Crane, James
  Cusack, Susan Leekam, Damian Milton, Jeremy~R Parr, and Elizabeth Pellicano.
  2019.
\newblock Making the future together: Shaping autism research through
  meaningful participation.
\newblock \emph{Autism}, 23(4):943--953.

\bibitem[{Floridi(2014)}]{floridi2014open}
Floridi, Luciano. 2014.
\newblock Open data, data protection, and group privacy.
\newblock \emph{Philosophy \& Technology}, 27(1):1--3.

\bibitem[{Fort, Adda, and Cohen(2011)}]{fort-etal-2011-last}
Fort, Kar{\"e}n, Gilles Adda, and K.~Bretonnel Cohen. 2011.
\newblock Last words: {A}mazon {M}echanical {T}urk: Gold mine or coal mine?
\newblock \emph{Computational Linguistics}, 37(2):413--420.

\bibitem[{Fraser et~al.(2019)Fraser, Zeller, Smith, Mohammad, and
  Rudzicz}]{fraser-etal-2019-feel}
Fraser, Kathleen~C., Frauke Zeller, David~Harris Smith, Saif~M. Mohammad, and
  Frank Rudzicz. 2019.
\newblock How do we feel when a robot dies? emotions expressed on {T}witter
  before and after hitch{BOT}{'}s destruction.
\newblock In \emph{Proceedings of the Tenth Workshop on Computational
  Approaches to Subjectivity, Sentiment and Social Media Analysis}, pages
  62--71, Minneapolis, USA.

\bibitem[{Gallagher et~al.(2021)Gallagher, Frank, Mitchell, Schwartz, Reagan,
  Danforth, and Dodds}]{gallagher2021generalized}
Gallagher, Ryan~J, Morgan~R Frank, Lewis Mitchell, Aaron~J Schwartz, Andrew~J
  Reagan, Christopher~M Danforth, and Peter~Sheridan Dodds. 2021.
\newblock Generalized word shift graphs: A method for visualizing and
  explaining pairwise comparisons between texts.
\newblock \emph{EPJ Data Science}, 10(1):4.

\bibitem[{Gerras and Wong(2016)}]{gerras2016moving}
Gerras, Stephen~J and Leonard Wong. 2016.
\newblock Moving beyond the {MBTI}.
\newblock \emph{Military review}.

\bibitem[{Google(2018)}]{what_if_2018}
Google. 2018.
\newblock Google {AI} Blog.
  \url{https://ai.googleblog.com/2018/09/the-what-if-tool-code-free-probing-of.html}.

\bibitem[{Grant(2013)}]{grant_2013}
Grant, Adam. 2013.
\newblock Say goodbye to {MBTI}, the fad that won't die.
\newblock Linkedin.
  \url{https://www.linkedin.com/pulse/20130917155206-69244073-say-goodbye-to-mbti-the-fad-that-won-t-die}.

\bibitem[{Guntuku et~al.(2019)Guntuku, Schneider, Pelullo, Young, Wong, Ungar,
  Polsky, Volpp, and Merchant}]{guntuku2019studying}
Guntuku, Sharath~Chandra, Rachelle Schneider, Arthur Pelullo, Jami Young,
  Vivien Wong, Lyle Ungar, Daniel Polsky, Kevin~G Volpp, and Raina Merchant.
  2019.
\newblock Studying expressions of loneliness in individuals using {T}witter: an
  observational study.
\newblock \emph{BMJ open}, 9(11):e030355.

\bibitem[{Gururangan et~al.(2018)Gururangan, Swayamdipta, Levy, Schwartz,
  Bowman, and Smith}]{gururangan2018annotation}
Gururangan, Suchin, Swabha Swayamdipta, Omer Levy, Roy Schwartz, Samuel~R
  Bowman, and Noah~A Smith. 2018.
\newblock Annotation artifacts in natural language inference data.
\newblock \emph{arXiv preprint arXiv:1803.02324}.

\bibitem[{Hall(2014)}]{hall2014not}
Hall, Lisa. 2014.
\newblock ‘{W}ith’not ‘about’: {E}merging paradigms for research in a
  cross-cultural space.
\newblock \emph{International Journal of Research \& Method in Education},
  37(4):376--389.

\bibitem[{Harris(1954)}]{harris1954distributional}
Harris, Zellig~S. 1954.
\newblock Distributional structure.
\newblock \emph{Word}, 10(2-3):146--162.

\bibitem[{Hertzmann(2020)}]{hertzmann2020computers}
Hertzmann, Aaron. 2020.
\newblock Computers do not make art, people do.
\newblock \emph{Communications of the ACM}, 63(5):45--48.

\bibitem[{Hipson and Mohammad(2021)}]{hipson2021emotion}
Hipson, Will~E. and Saif~M. Mohammad. 2021.
\newblock Emotion dynamics in movie dialogues.
\newblock \emph{PLOS ONE}, 16:1--19.

\bibitem[{Hollenstein(2015)}]{hollenstein2015time}
Hollenstein, Tom. 2015.
\newblock This time, it’s real: Affective flexibility, time scales, feedback
  loops, and the regulation of emotion.
\newblock \emph{Emotion Review}, 7(4):308--315.

\bibitem[{Hovy and Yang(2021)}]{hovy-yang-2021-importance}
Hovy, Dirk and Diyi Yang. 2021.
\newblock The importance of modeling social factors of language: Theory and
  practice.
\newblock In \emph{Proceedings of the 2021 Conference of the North American
  Chapter of the Association for Computational Linguistics: Human Language
  Technologies}, pages 588--602, Association for Computational Linguistics,
  Online.

\bibitem[{Humphries, Mertens, and Truman(2020)}]{humphries2020arguments}
Humphries, Beth, Donna~M Mertens, and Carole Truman. 2020.
\newblock Arguments for an ‘emancipatory’research paradigm.
\newblock In \emph{Research and inequality}. Routledge, pages 3--23.

\bibitem[{Irani and Silberman(2013)}]{irani2013turkopticon}
Irani, Lilly~C and M~Six Silberman. 2013.
\newblock Turkopticon: Interrupting worker invisibility in {A}mazon
  {M}echanical {T}urk.
\newblock In \emph{Proceedings of the SIGCHI conference on human factors in
  computing systems}, pages 611--620.

\bibitem[{Johnson(2019)}]{johnson_2019}
Johnson, Khari. 2019.
\newblock How {AI} companies can avoid ethics washing.
\newblock VentureBeat.
  \url{https://venturebeat.com/2019/07/17/how-ai-companies-can-avoid-ethics-washing/}.

\bibitem[{Kalluri(2020)}]{kalluri2020don}
Kalluri, Pratyusha. 2020.
\newblock Don't ask if {A}rtificial {I}ntelligence is good or fair, ask how it
  shifts power.
\newblock \emph{Nature}, 583(7815):169--169.

\bibitem[{Karam et~al.(2014)Karam, Provost, Singh, Montgomery, Archer,
  Harrington, and Mcinnis}]{karam2014ecologically}
Karam, Zahi~N, Emily~Mower Provost, Satinder Singh, Jennifer Montgomery,
  Christopher Archer, Gloria Harrington, and Melvin~G Mcinnis. 2014.
\newblock Ecologically valid long-term mood monitoring of individuals with
  bipolar disorder using speech.
\newblock In \emph{2014 IEEE international conference on acoustics, speech and
  signal processing (ICASSP)}, pages 4858--4862, IEEE.

\bibitem[{Keyes(2019)}]{keyes_2019}
Keyes, Os. 2019.
\newblock Counting the countless.
\newblock REAL LIFE. \url{https://reallifemag.com/counting-the-countless/}.

\bibitem[{Kiritchenko et~al.(2020)Kiritchenko, Hipson, Coplan, and
  Mohammad}]{kiritchenko-etal-2020-solo}
Kiritchenko, Svetlana, Will Hipson, Robert Coplan, and Saif~M. Mohammad. 2020.
\newblock {SOLO}: A corpus of tweets for examining the state of being alone.
\newblock In \emph{Proceedings of the 12th Language Resources and Evaluation
  Conference}, pages 1567--1577, Marseille, France.

\bibitem[{Kiritchenko and Mohammad(2018)}]{kiritchenko-mohammad-2018-examining}
Kiritchenko, Svetlana and Saif Mohammad. 2018.
\newblock Examining gender and race bias in two hundred sentiment analysis
  systems.
\newblock In \emph{Proceedings of the Seventh Joint Conference on Lexical and
  Computational Semantics}, pages 43--53, New Orleans, Louisiana.

\bibitem[{Klenner et~al.(2020)Klenner, G{\"o}hring, Amsler, Ebling, Tuggener,
  H{\"u}rlimann, and Volk}]{klenner2020harmonization}
Klenner, Manfred, Anne G{\"o}hring, Michael Amsler, Sarah Ebling, Don Tuggener,
  Manuela H{\"u}rlimann, and Martin Volk. 2020.
\newblock Harmonization sometimes harms.
\newblock In \emph{Proceedings of the 5th Swiss Text Analytics Conference
  (SwissText) \& 16th Conference on Natural Language Processing (KONVENS)},
  Winterthur.

\bibitem[{Kucher, Paradis, and Kerren(2018)}]{kucher2018visual}
Kucher, Kostiantyn, Carita Paradis, and Andreas Kerren. 2018.
\newblock Visual analysis of sentiment and stance in social media texts.
\newblock In \emph{EuroVis (Posters)}, pages 49--51.

\bibitem[{Lakoff(2008)}]{lakoff2008women}
Lakoff, George. 2008.
\newblock \emph{Women, fire, and dangerous things: What categories reveal about
  the mind}.
\newblock University of Chicago press.

\bibitem[{Lazarus(1991)}]{lazarus1991progress}
Lazarus, Richard~S. 1991.
\newblock Progress on a cognitive-motivational-relational theory of emotion.
\newblock \emph{American psychologist}, 46(8):819.

\bibitem[{Lindsey(2015)}]{lindsey2015sociology}
Lindsey, Linda~L. 2015.
\newblock The sociology of gender theoretical perspectives and feminist
  frameworks.
\newblock In \emph{Gender roles}. Routledge, pages 23--48.

\bibitem[{Luo et~al.(2021)Luo, Ivison, Han, and Poon}]{luo2021local}
Luo, Siwen, Hamish Ivison, Caren Han, and Josiah Poon. 2021.
\newblock Local interpretations for explainable natural language processing: A
  survey.
\newblock \emph{arXiv preprint arXiv:2103.11072}.

\bibitem[{Lysaght et~al.(2019)Lysaght, Lim, Xafis, and Ngiam}]{lysaght2019ai}
Lysaght, Tamra, Hannah~Yeefen Lim, Vicki Xafis, and Kee~Yuan Ngiam. 2019.
\newblock {AI}-assisted decision-making in healthcare.
\newblock \emph{Asian Bioethics Review}, 11(3):299--314.

\bibitem[{MacAvaney et~al.(2021)MacAvaney, Mittu, Coppersmith, Leintz, and
  Resnik}]{macavaney-etal-2021-community}
MacAvaney, Sean, Anjali Mittu, Glen Coppersmith, Jeff Leintz, and Philip
  Resnik. 2021.
\newblock Community-level research on suicidality prediction in a secure
  environment: Overview of the {CLP}sych 2021 shared task.
\newblock In \emph{Proceedings of the Seventh Workshop on Computational
  Linguistics and Clinical Psychology: Improving Access}, pages 70--80,
  Association for Computational Linguistics, Online.

\bibitem[{McStay(2020)}]{mcstay2020emotional}
McStay, Andrew. 2020.
\newblock Emotional {AI}, soft biometrics and the surveillance of emotional
  life: An unusual consensus on privacy.
\newblock \emph{Big Data \& Society}, 7(1):2053951720904386.

\bibitem[{Mitchell et~al.(2019)Mitchell, Wu, Zaldivar, Barnes, Vasserman,
  Hutchinson, Spitzer, Raji, and Gebru}]{mitchell2019model}
Mitchell, Margaret, Simone Wu, Andrew Zaldivar, Parker Barnes, Lucy Vasserman,
  Ben Hutchinson, Elena Spitzer, Inioluwa~Deborah Raji, and Timnit Gebru. 2019.
\newblock Model cards for model reporting.
\newblock In \emph{Proceedings of the conference on fairness, accountability,
  and transparency}, pages 220--229.

\bibitem[{Mohammad(2011)}]{mohammad-2011-upon}
Mohammad, Saif. 2011.
\newblock From once upon a time to happily ever after: Tracking emotions in
  novels and fairy tales.
\newblock In \emph{Proceedings of the 5th {ACL}-{HLT} Workshop on Language
  Technology for Cultural Heritage, Social Sciences, and Humanities}, pages
  105--114, Portland, OR, USA.

\bibitem[{Mohammad(2012)}]{mohammad-2012-emotional}
Mohammad, Saif. 2012.
\newblock {\#E}motional tweets.
\newblock In \emph{*{SEM} 2012: The First Joint Conference on Lexical and
  Computational Semantics {--} Volume 1: Proceedings of the main conference and
  the shared task, and Volume 2: Proceedings of the Sixth International
  Workshop on Semantic Evaluation ({S}em{E}val 2012)}, pages 246--255,
  Montr{\'e}al, Canada.

\bibitem[{Mohammad et~al.(2018)Mohammad, Bravo-Marquez, Salameh, and
  Kiritchenko}]{mohammad-etal-2018-semeval}
Mohammad, Saif, Felipe Bravo-Marquez, Mohammad Salameh, and Svetlana
  Kiritchenko. 2018.
\newblock {S}em{E}val-2018 task 1: Affect in tweets.
\newblock In \emph{Proceedings of The 12th International Workshop on Semantic
  Evaluation}, pages 1--17, New Orleans, Louisiana.

\bibitem[{Mohammad and
  Kiritchenko(2018)}]{mohammad-kiritchenko-2018-understanding}
Mohammad, Saif and Svetlana Kiritchenko. 2018.
\newblock Understanding emotions: A dataset of tweets to study interactions
  between affect categories.
\newblock In \emph{Proceedings of the Eleventh International Conference on
  Language Resources and Evaluation ({LREC} 2018)}, Miyazaki, Japan.

\bibitem[{Mohammad(2018)}]{vad-acl2018}
Mohammad, Saif~M. 2018.
\newblock Obtaining reliable human ratings of valence, arousal, and dominance
  for 20,000 {E}nglish words.
\newblock In \emph{Proceedings of The Annual Conference of the Association for
  Computational Linguistics (ACL)}, Melbourne, Australia.

\bibitem[{Mohammad(2020)}]{mohammad2020practical}
Mohammad, Saif~M. 2020.
\newblock Practical and ethical considerations in the effective use of emotion
  and sentiment lexicons.
\newblock \emph{arXiv:2011.03492}.

\bibitem[{Mohammad(2021{\natexlab{a}})}]{mohammad2021ethics}
Mohammad, Saif~M. 2021{\natexlab{a}}.
\newblock Ethics sheets for {AI} tasks.
\newblock In \emph{Proceedings of the 60th Annual Meeting of the Association
  for Computational Linguistics}, Dublin, Ireland.

\bibitem[{Mohammad(2021{\natexlab{b}})}]{mohammad2020survey}
Mohammad, Saif~M. 2021{\natexlab{b}}.
\newblock Sentiment analysis: Automatically detecting valence, emotions, and
  other affectual states from text.
\newblock In Herbert~L. Meiselman, editor, \emph{Emotion Measurement (Second
  Edition)}, second edition edition. Woodhead Publishing, pages 323--379.

\bibitem[{Mohammad et~al.(2016)Mohammad, Kiritchenko, Sobhani, Zhu, and
  Cherry}]{StanceSemEval2016}
Mohammad, Saif~M., Svetlana Kiritchenko, Parinaz Sobhani, Xiaodan Zhu, and
  Colin Cherry. 2016.
\newblock Semeval-2016 task 6: Detecting stance in tweets.
\newblock In \emph{Proceedings of the International Workshop on Semantic
  Evaluation}, SemEval '16, San Diego, California.

\bibitem[{Mohammad, Sobhani, and Kiritchenko(2017)}]{MohammadSK17}
Mohammad, Saif~M., Parinaz Sobhani, and Svetlana Kiritchenko. 2017.
\newblock Stance and sentiment in tweets.
\newblock \emph{Special Section of the ACM Transactions on Internet Technology
  on Argumentation in Social Media}, 17(3):1--23.

\bibitem[{Mohammad and Turney(2013)}]{Mohammad13}
Mohammad, Saif~M. and Peter~D. Turney. 2013.
\newblock Crowdsourcing a word-emotion association lexicon.
\newblock \emph{Computational Intelligence}, 29(3):436--465.

\bibitem[{Monteiro(2019)}]{monteiro2019ruined}
Monteiro, Mike. 2019.
\newblock \emph{Ruined by design: How designers destroyed the world, and what
  we can do to fix it}.
\newblock Mule Design.

\bibitem[{Motti and Evmenova(2020)}]{10.1007/978-3-030-25629-6_42}
Motti, Vivian~Genaro and Anna Evmenova. 2020.
\newblock Designing technologies for neurodiverse users: Considerations from
  research practice.
\newblock In \emph{Human Interaction and Emerging Technologies}, pages
  268--274, Springer International Publishing, Cham.

\bibitem[{Mozafari, Weiger, and Hammerschmidt(2020)}]{mozafari2020chatbot}
Mozafari, Nika, Welf~H Weiger, and Maik Hammerschmidt. 2020.
\newblock The chatbot disclosure dilemma: Desirable and undesirable effects of
  disclosing the non-human identity of chatbots.
\newblock In \emph{ICIS}, pages 1--18.

\bibitem[{Mulligan, Kluttz, and Kohli(2019)}]{mulligan2019shaping}
Mulligan, Deirdre~K, Daniel Kluttz, and Nitin Kohli. 2019.
\newblock Shaping our tools: Contestability as a means to promote responsible
  algorithmic decision making in the professions.
\newblock \emph{Available at SSRN 3311894}.

\bibitem[{Nielsen(2011)}]{nielsen2011new}
Nielsen, Finn~{\AA}rup. 2011.
\newblock A new {ANEW}: Evaluation of a word list for sentiment analysis in
  microblogs.
\newblock In \emph{Proceedings of the ESWC Workshop on `Making Sense of
  Microposts': Big things come in small packages}, pages 93--98, Heraklion,
  Crete.

\bibitem[{Noel(2016)}]{noel2016promoting}
Noel, Lesley-Ann. 2016.
\newblock Promoting an emancipatory research paradigm in design education and
  practice.
\newblock In \emph{Proceedings of DRS2016 International Conference, Vol. 6:
  Future–Focused Thinking}, pages 27--30, Brighton, United Kingdom.

\bibitem[{Oliver(1997)}]{oliver1997emancipatory}
Oliver, Michael. 1997.
\newblock Emancipatory research: Realistic goal or impossible dream.
\newblock \emph{Doing disability research}, 2:15--31.

\bibitem[{Osgood, Suci, and Tannenbaum(1957)}]{osgood1957measurement}
Osgood, Charles~Egerton, George~J Suci, and Percy~H Tannenbaum. 1957.
\newblock \emph{The measurement of meaning}.
\newblock 47. University of Illinois press.

\bibitem[{Panesar(2019)}]{panesar2019machine}
Panesar, Arjun. 2019.
\newblock \emph{Machine learning and {AI} for healthcare}.
\newblock Springer.

\bibitem[{Pang, Lee, and Vaithyanathan(2002)}]{pang-etal-2002-thumbs}
Pang, Bo, Lillian Lee, and Shivakumar Vaithyanathan. 2002.
\newblock Thumbs up? sentiment classification using machine learning
  techniques.
\newblock In \emph{Proceedings of the 2002 Conference on Empirical Methods in
  Natural Language Processing ({EMNLP} 2002)}, pages 79--86.

\bibitem[{Paul and Dredze(2011)}]{paul2011you}
Paul, Michael~J and Mark Dredze. 2011.
\newblock You are what you tweet: Analyzing {T}witter for public health.
\newblock \emph{Proceedings of the Fifth international AAAI conference on
  weblogs and social media}, 5(1):265--272.

\bibitem[{Perez(2019)}]{perez2019invisible}
Perez, Caroline~Criado. 2019.
\newblock \emph{Invisible women: Exposing data bias in a world designed for
  men}.
\newblock Random House.

\bibitem[{Picard(2000)}]{picard2000affective}
Picard, Rosalind~W. 2000.
\newblock \emph{Affective computing}.
\newblock MIT press.

\bibitem[{Pinker(2007)}]{pinker2007stuff}
Pinker, Steven. 2007.
\newblock \emph{The stuff of thought: Language as a window into human nature}.
\newblock Penguin.

\bibitem[{Poliak et~al.(2018)Poliak, Naradowsky, Haldar, Rudinger, and
  Van~Durme}]{poliak-etal-2018-hypothesis}
Poliak, Adam, Jason Naradowsky, Aparajita Haldar, Rachel Rudinger, and Benjamin
  Van~Durme. 2018.
\newblock Hypothesis only baselines in natural language inference.
\newblock In \emph{Proceedings of the Seventh Joint Conference on Lexical and
  Computational Semantics}, pages 180--191, New Orleans, Louisiana.

\bibitem[{Purdie-Vaughns and Eibach(2008)}]{purdie2008intersectional}
Purdie-Vaughns, Valerie and Richard~P Eibach. 2008.
\newblock Intersectional invisibility: The distinctive advantages and
  disadvantages of multiple subordinate-group identities.
\newblock \emph{Sex roles}, 59(5):377--391.

\bibitem[{Purver and Battersby(2012)}]{purver2012experimenting}
Purver, Matthew and Stuart Battersby. 2012.
\newblock Experimenting with distant supervision for emotion classification.
\newblock In \emph{Proceedings of the 13th Conference of the European Chapter
  of the Association for Computational Linguistics}, pages 482--491.

\bibitem[{Quercia et~al.(2012)Quercia, Ellis, Capra, and
  Crowcroft}]{10.1145/2145204.2145347}
Quercia, Daniele, Jonathan Ellis, Licia Capra, and Jon Crowcroft. 2012.
\newblock Tracking "{G}ross community happiness" from tweets.
\newblock In \emph{Proceedings of the ACM 2012 Conference on Computer Supported
  Cooperative Work}, CSCW '12, page 965–968, Association for Computing
  Machinery, New York, NY, USA.

\bibitem[{Resnik et~al.(2015)Resnik, Armstrong, Claudino, Nguyen, Nguyen, and
  Boyd-Graber}]{resnik-etal-2015-beyond}
Resnik, Philip, William Armstrong, Leonardo Claudino, Thang Nguyen, Viet-An
  Nguyen, and Jordan Boyd-Graber. 2015.
\newblock Beyond {LDA}: Exploring supervised topic modeling for
  depression-related language in {T}witter.
\newblock In \emph{Proceedings of the 2nd Workshop on Computational Linguistics
  and Clinical Psychology: From Linguistic Signal to Clinical Reality}, pages
  99--107, Denver, Colorado.

\bibitem[{Rosenthal et~al.(2015)Rosenthal, Nakov, Kiritchenko, Mohammad,
  Ritter, and Stoyanov}]{rosenthal-etal-2015-semeval}
Rosenthal, Sara, Preslav Nakov, Svetlana Kiritchenko, Saif Mohammad, Alan
  Ritter, and Veselin Stoyanov. 2015.
\newblock {S}em{E}val-2015 task 10: Sentiment analysis in {T}witter.
\newblock In \emph{Proceedings of the 9th International Workshop on Semantic
  Evaluation ({S}em{E}val 2015)}, pages 451--463, Denver, Colorado.

\bibitem[{Rosenthal et~al.(2014)Rosenthal, Ritter, Nakov, and
  Stoyanov}]{rosenthal-etal-2014-semeval}
Rosenthal, Sara, Alan Ritter, Preslav Nakov, and Veselin Stoyanov. 2014.
\newblock {S}em{E}val-2014 task 9: Sentiment analysis in {T}witter.
\newblock In \emph{Proceedings of the 8th International Workshop on Semantic
  Evaluation ({S}em{E}val 2014)}, pages 73--80, Association for Computational
  Linguistics, Dublin, Ireland.

\bibitem[{R{\"o}ttger et~al.(2020)R{\"o}ttger, Vidgen, Nguyen, Waseem,
  Margetts, and Pierrehumbert}]{rottger2020hatecheck}
R{\"o}ttger, Paul, Bertram Vidgen, Dong Nguyen, Zeerak Waseem, Helen Margetts,
  and Janet Pierrehumbert. 2020.
\newblock Hatecheck: Functional tests for hate speech detection models.
\newblock \emph{arXiv preprint arXiv:2012.15606}.

\bibitem[{Ruder(2020)}]{ruder_2020}
Ruder, Sebastian Ruder~Sebastian. 2020.
\newblock Why you should do nlp beyond {E}nglish.
\newblock \url{https://ruder.io/nlp-beyond-english/index.html}.

\bibitem[{Russell(1980)}]{russell1980circumplex}
Russell, James~A. 1980.
\newblock A circumplex model of affect.
\newblock \emph{Journal of personality and social psychology}, 39(6):1161.

\bibitem[{Russell(2003)}]{russell2003core}
Russell, James~A. 2003.
\newblock Core affect and the psychological construction of emotion.
\newblock \emph{Psychological review}, 110(1):145.

\bibitem[{Russell and Mehrabian(1977)}]{russell1977evidence}
Russell, James~A and Albert Mehrabian. 1977.
\newblock Evidence for a three-factor theory of emotions.
\newblock \emph{Journal of research in Personality}, 11(3):273--294.

\bibitem[{Schaar(2010)}]{schaar2010privacy}
Schaar, Peter. 2010.
\newblock Privacy by design.
\newblock \emph{Identity in the Information Society}, 3(2):267--274.

\bibitem[{Scherer(1999)}]{scherer1999appraisal}
Scherer, Klaus~R. 1999.
\newblock \emph{Appraisal theory.}
\newblock John Wiley \& Sons Ltd.

\bibitem[{Schwartz et~al.(2013)Schwartz, Eichstaedt, Kern, Dziurzynski, Lucas,
  Agrawal, Park, Lakshmikanth, Jha, Seligman
  et~al.}]{schwartz2013characterizing}
Schwartz, Hansen~Andrew, Johannes~C Eichstaedt, Margaret~L Kern, Lukasz
  Dziurzynski, Richard~E Lucas, Megha Agrawal, Gregory~J Park, Shrinidhi~K
  Lakshmikanth, Sneha Jha, Martin~EP Seligman, et~al. 2013.
\newblock Characterizing geographic variation in well-being using tweets.
\newblock In \emph{Seventh International AAAI Conference on Weblogs and Social
  Media}, pages 583--591.

\bibitem[{Schwartz et~al.(2020)Schwartz, Dodge, Smith, and
  Etzioni}]{schwartz2020green}
Schwartz, Roy, Jesse Dodge, Noah~A Smith, and Oren Etzioni. 2020.
\newblock Green {AI}.
\newblock \emph{Communications of the ACM}, 63(12):54--63.

\bibitem[{Seale et~al.(2015)Seale, Nind, Tilley, and
  Chapman}]{seale2015negotiating}
Seale, Jane, Melanie Nind, Liz Tilley, and Rohhss Chapman. 2015.
\newblock Negotiating a third space for participatory research with people with
  learning disabilities: An examination of boundaries and spatial practices.
\newblock \emph{Innovation: The European Journal of Social Science Research},
  28(4):483--497.

\bibitem[{Shmueli et~al.(2021)Shmueli, Fell, Ray, and Ku}]{shmueli2021beyond}
Shmueli, Boaz, Jan Fell, Soumya Ray, and Lun-Wei Ku. 2021.
\newblock Beyond fair pay: Ethical implications of nlp crowdsourcing.
\newblock \emph{arXiv preprint arXiv:2104.10097}.

\bibitem[{Snow(2020)}]{snow_2020}
Snow, Shane. 2020.
\newblock That personality test may be discriminating people... and making your
  company dumber.
\newblock Linkedin.
  \url{https://www.linkedin.com/pulse/personality-test-may-discriminating-people-making-your-shane-snow}.

\bibitem[{Soleymani et~al.(2017)Soleymani, Garcia, Jou, Schuller, Chang, and
  Pantic}]{soleymani2017survey}
Soleymani, Mohammad, David Garcia, Brendan Jou, Bj{\"o}rn Schuller, Shih-Fu
  Chang, and Maja Pantic. 2017.
\newblock A survey of multimodal sentiment analysis.
\newblock \emph{Image and Vision Computing}, 65:3--14.

\bibitem[{Spinuzzi(2005)}]{spinuzzi2005methodology}
Spinuzzi, Clay. 2005.
\newblock The methodology of participatory design.
\newblock \emph{Technical communication}, 52(2):163--174.

\bibitem[{Standing and Standing(2018)}]{standing2018ethical}
Standing, Susan and Craig Standing. 2018.
\newblock The ethical use of crowdsourcing.
\newblock \emph{Business Ethics: A European Review}, 27(1):72--80.

\bibitem[{Stone and Priestley(1996)}]{stone1996parasites}
Stone, Emma and Mark Priestley. 1996.
\newblock Parasites, pawns and partners: Disability research and the role of
  non-disabled researchers.
\newblock \emph{British journal of sociology}, pages 699--716.

\bibitem[{Strubell, Ganesh, and McCallum(2020)}]{Strubell_Ganesh_McCallum_2020}
Strubell, Emma, Ananya Ganesh, and Andrew McCallum. 2020.
\newblock Energy and policy considerations for modern deep learning research.
\newblock \emph{Proceedings of the AAAI Conference on Artificial Intelligence},
  34(09):13693--13696.

\bibitem[{Tausczik and Pennebaker(2010)}]{tausczik2010psychological}
Tausczik, Yla~R and James~W Pennebaker. 2010.
\newblock The psychological meaning of words: {LIWC} and computerized text
  analysis methods.
\newblock \emph{Journal of language and social psychology}, 29(1):24--54.

\bibitem[{Thaine and Penn(2021)}]{thaine-penn-2021-chinese}
Thaine, Patricia and Gerald Penn. 2021.
\newblock The {C}hinese remainder theorem for compact, task-precise, efficient
  and secure word embeddings.
\newblock In \emph{Proceedings of the 16th Conference of the European Chapter
  of the Association for Computational Linguistics: Main Volume}, pages
  3512--3521, Online.

\bibitem[{Trewin et~al.(2019)Trewin, Basson, Muller, Branham, Treviranus,
  Gruen, Hebert, Lyckowski, and Manser}]{trewin2019considerations}
Trewin, Shari, Sara Basson, Michael Muller, Stacy Branham, Jutta Treviranus,
  Daniel Gruen, Daniel Hebert, Natalia Lyckowski, and Erich Manser. 2019.
\newblock Considerations for {AI} fairness for people with disabilities.
\newblock \emph{AI Matters}, 5(3):40--63.

\bibitem[{Turney(2002)}]{turney-2002-thumbs}
Turney, Peter. 2002.
\newblock Thumbs up or thumbs down? semantic orientation applied to
  unsupervised classification of reviews.
\newblock In \emph{Proceedings of the 40th Annual Meeting of the Association
  for Computational Linguistics}, pages 417--424, Philadelphia, Pennsylvania,
  USA.

\bibitem[{Wakefield(2021)}]{wakefield_2021}
Wakefield, Jane. 2021.
\newblock {AI} emotion-detection software tested on {U}yghurs.
\newblock BBC. \url{https://www.bbc.com/news/technology-57101248}.

\bibitem[{Wiebe, Wilson, and Cardie(2005)}]{wiebe2005annotating}
Wiebe, Janyce, Theresa Wilson, and Claire Cardie. 2005.
\newblock Annotating expressions of opinions and emotions in language.
\newblock \emph{Language resources and evaluation}, 39(2):165--210.

\bibitem[{Winkler et~al.(2019)Winkler, Fink, Toberer, Enk, Deinlein,
  Hofmann-Wellenhof, Thomas, Lallas, Blum, Stolz
  et~al.}]{winkler2019association}
Winkler, Julia~K, Christine Fink, Ferdinand Toberer, Alexander Enk, Teresa
  Deinlein, Rainer Hofmann-Wellenhof, Luc Thomas, Aimilios Lallas, Andreas
  Blum, Wilhelm Stolz, et~al. 2019.
\newblock Association between surgical skin markings in dermoscopic images and
  diagnostic performance of a deep learning convolutional neural network for
  melanoma recognition.
\newblock \emph{JAMA dermatology}, 155(10):1135--1141.

\bibitem[{Woensel and Nevil(2019)}]{woensel_nevil_2019}
Woensel, Lieve~Van and Nissy Nevil. 2019.
\newblock What if your emotions were tracked to spy on you?
\newblock European Parliamentary Research Service, PE 634.415.
  \url{https://www.europarl.europa.eu/RegData/etudes/ATAG/2019/634415/EPRS_ATA(2019)634415_EN.pdf}.

\bibitem[{Yu, Beam, and Kohane(2018)}]{yu2018artificial}
Yu, Kun-Hsing, Andrew~L Beam, and Isaac~S Kohane. 2018.
\newblock Artificial {I}ntelligence in healthcare.
\newblock \emph{Nature biomedical engineering}, 2(10):719--731.

\bibitem[{Zhang, Wang, and Liu(2018)}]{zhang2018deep}
Zhang, Lei, Shuai Wang, and Bing Liu. 2018.
\newblock Deep learning for sentiment analysis: A survey.
\newblock \emph{Wiley Interdisciplinary Reviews: Data Mining and Knowledge
  Discovery}, 8(4):e1253.

\end{thebibliography}

\end{document}